\documentclass[11pt, a4paper]{article}

\usepackage[utf8]{inputenc}
\usepackage[T1]{fontenc}
\usepackage{amsmath, amssymb, amsfonts}
\usepackage{graphicx} 
\usepackage{booktabs} 
\usepackage{multirow} 
\usepackage{geometry}
\usepackage{hyperref} 
\usepackage{authblk}  
\usepackage{float}
\usepackage{caption}
\usepackage{url}      
\usepackage{float}
\usepackage[symbol]{footmisc}
\usepackage{authblk}
\usepackage{caption}
\usepackage{etoolbox}
\AtBeginEnvironment{table}{\small}
\hypersetup{
    colorlinks=true,
    linkcolor=blue,
    filecolor=magenta,      
    urlcolor=cyan,
    citecolor=blue,
}

\title{\textbf{Quasi-multimodal-based pathophysiological feature learning for retinal disease diagnosis}}

\author[1]{Lu Zhang}
\author[1]{Huizhen Yu}
\author[2,3]{Zuowei Wang}
\author[5]{Fu Gui}
\author[2]{Yatu Guo$^{*}$}
\author[2,3,4]{Wei Zhang$^{*}$}
\author[1]{Mengyu Jia$^{*}$}

\affil[1]{College of Precision Instruments and Optoelectronics Engineering, Tianjin University, Tianjin 300072, China}
\affil[2]{Tianjin Key Laboratory of Ophthalmology and Visual Science, Tianjin Eye Institute, Tianjin Eye Hospital, Tianjin 300020, China}
\affil[3]{Clinical College of Ophthalmology, Tianjin Medical University, Tianjin 300020, China}
\affil[4]{Nankai University Affiliated Eye Hospital, Nankai University, Tianjin 300020, China}
\affil[5]{Department of Ophthalmology, The Second Affiliated Hospital of Nanchang University, Nanchang 330006, China}

\begin{document}
\date{}
\maketitle
\footnotetext[1]{Corresponding  authors.}

\vspace{-3em}

\begin{abstract}
Retinal diseases spanning a broad spectrum can be effectively identified and diagnosed using complementary signals from multimodal data. However, multimodal diagnosis in ophthalmic practice is typically challenged in terms of data heterogeneity, potential invasiveness, registration complexity, and so on. As such, a unified framework that integrates multimodal data synthesis and fusion is proposed for retinal disease classification and grading. Specifically, the synthesized multimodal data incorporates fundus fluorescein angiography (FFA), multispectral imaging (MSI), and saliency maps that emphasize latent lesions as well as optic disc/cup regions. Parallel models are independently trained to learn modality-specific representations that capture cross-pathophysiological signatures. These features are then adaptively calibrated within and across modalities to perform information pruning and flexible integration according to downstream tasks. The proposed learning system is thoroughly interpreted through visualizations in both image and feature spaces. Extensive experiments on two public datasets demonstrated the superiority of our approach over state-of-the-art ones in the tasks of multi-label classification (F1-score: 0.683, AUC: 0.953) and diabetic retinopathy grading (Accuracy:0.842, Kappa: 0.861). This work not only enhances the accuracy and efficiency of retinal disease screening but also offers a scalable framework for data augmentation across various medical imaging modalities.
\end{abstract}

\vspace{1em}

\newpage
\section{Introduction}
The unique anatomy and transparency of the eye allow the retina to serve as a window for noninvasive, high-resolution imaging, providing insights not only on retinal diseases but also on neurological disorders and systemic microcirculation \cite{Kashani2021}. Retinal imaging, with its capability for early detection, diagnosis, and disease management, is a powerful and intuitive tool for comprehensive health assessment \cite{Lim2020}\cite{Shi2024}. Among various retinal imaging modalities, color fundus photography (CFP) is a well-established technique, known for its non-invasive nature, high spatiotemporal resolution, and effectiveness in diagnosing common retinal diseases \cite{Ong2004}\cite{Mokwa2013}\cite{Sun2022}. Fundus fluorescein angiography (FFA), which is implemented by injecting sodium fluorescein, is considered the gold standard for evaluating functional state of retinal circulation, enabling highly sensitive detection on non-perfusion areas, vascular leakage, microaneurysms, and so on \cite{Wang2017a}\cite{Pan2020}\cite{Nanegrungsunk2022} \cite{Shen2024}. Multispectral imaging (MSI) leverages tissue-specific absorption across the spectrum to reveal subtle biochemical and structural changes that may not be detectable with either CFP or FFA \cite{Ma2023}, e.g., Diabetic Neuropathy (DN), Retinal Vein Occlusion (RVO), and Age-related macular degeneration (AMD) \cite{Li2015}\cite{deCarvalho2020}. 

While each imaging technique offers unique advantages, no single modality can encompass the full range of retinal diseases, which essentially underscores the need for multimodal imaging. However, physical deployment of multimodal imaging is impeded by both internal and external factors. The internal ones include, but are not limited to, the invasive nature and potential risks (e.g., FFA), spatial misalignment due to rapid saccadic eye movements during inspection (e.g., MSI), and lack of standardized protocols pertained to each modality \cite{Dalmaz2022}\cite{Zhang2022}\cite{Shen2024}. The external ones involve obstacles in alignment, fusion, and the comprehensive interpretation of either raw data or learned features across multiple modalities \cite{Jain2021}\cite{Sindel2022}\cite{ChenHong2024}. 
   
Given the dilemma above, quasi-multimodal imaging technique, which synchronizes images across various modalities, has been widely recognized as a promising approach for enhancing disease identification and diagnosis \cite{Sharma2019}\cite{Dalmaz2022}\cite{Ting2023}. By synthesizing arteriovenous-phase FFA and further combining it with the baseline CFP images, the diagnostic accuracy for multiple retinal diseases has been significantly improved \cite{Li2019a}\cite{Li2020b}. Similar approaches can be found outside ophthalmology as well. For instance, Pan et al. employed magnetic resonance imaging (MRI) to predict corresponding positron emission tomography (PET) images, which were then jointly used to detect neurological disorders \cite{Pan2021}. MSI synthesization has also been widely applied in fields such as remote sensing \cite{Rout2020}, biomedical imaging \cite{Ma2021}, and food engineering \cite{Dong2025}. Despite the extensive exploration of quasi-multimodal applications, challenges remain. i) Limited representation for specific modalities. For instance, arterial-phase FFA, which is more clinically informative than arteriovenous-phase FFA, has yet to be synthesized, mainly due to the subtle nature of arterial structures relative to the entire fundus. ii) Effective handling of information redundancy. Improper information tailing and fusion could adversely affect the diagnostic performance. iii) Reduced diagnostic in generalizability and robustness. This is arguably due to the gap between the fine-grained pixel-wise synthesization and the relatively less complicated downstream tasks. 

In this paper, a unified learning-based framework is proposed for exploring quasi-multimodal features, aimed at retinal disease classification and grading. The framework is implemented in dual stages. In Stage I, parallel models are independently trained to learn the mapping from CFP to various modalities that focus on pathophysiological signatures, including MSI, FFA (in both arterial and arteriovenous phases), and saliency maps that highlight latent lesion and optic-disc/cup regions in CFP. The learned representations are then transferred to Stage II in the form of high-level features. These features are first refined through task-driven finetuning of the corresponding encoders, guided jointly by task-specific objectives and fidelity regularization. Subsequently, they are adaptively calibrated both within and across modalities to reduce redundancy and support flexible, informative feature integration.
Our work makes the following key contributions: 

\begin{enumerate}
    \item We develop a unified architecture that integrates multimodal data synthesis and fusion for retinal disease classification and grading. To emphasize disease sensitivity in modality-specific synthesis, synthesis is finetuned jointly with the fusion pipeline in a task-driven manner. To the best of our knowledge, this is the first quasi-multimodal model that involves arterial-phase FFA and MSI.
    \item We propose a Cross-Pathophysiology Attention Module (CPAM) to adaptively prune and fuse multimodal representations. CPAM hierarchically calibrates features through a coupled attention mechanism comprising both modal-specific and modal-dependent components, enabling fine-grained, multi-scale alignment of pathophysiological features across modalities.
    \item Experiments on multi-label classification and diabetic retinopathy (DR) grading demonstrate the superiority of the proposed approach over selected state-of-the-art ones. Furthermore, extensive evaluations were conducted to interpret the contributions of each modality and calibration node, as well as the effectiveness of the adaptive feature calibration mechanism.
\end{enumerate}

\section{Related work}

\subsection{Automated diagnosis of retinal diseases via CFP}
In ophthalmology, deep learning has been proven effective in extracting meaningful features from large-scale CFP images. For example, Wang et al. developed a deep multi-task model for CFP-based DR grading representations \cite{Wang2021a}. The model was trained on the databases for both lesion segmentation and DR grading to collaboratively learn the disease-specific and disease-dependent. Sreng et al. proposed a two-stage learning system for effective glaucoma screening \cite{Sreng2020}. The first stage contained segmentation for the optic disc (OD) region by employing the DeepLabv3+, followed by localized prediction within the segmented area. Araújo et al. proposed a DR grading system that can support its decision by providing a grade uncertainty parameter \cite{Araujo2020}. The network architecture consisted of convolutional blocks that generated lesion maps indicative of lesion presence. To address the challenges of class imbalance and outliers commonly encountered in multi-label ocular disease classification, Luo et al. proposed a hybrid loss function combining focal loss and correntropy-induced loss \cite{Luo2021}. Yang and Yi developed a learning framework that incorporated data preprocessing, feature extraction via the DSRA-CNN, and disease classification, for CFP-based identification of multiple ocular diseases \cite{Yang2022}.

For all the above methods, promising diagnostic accuracy could be achieved given a sufficiently large and relevant training dataset. However, the challenge is how to quantify "sufficiency" and "relevance" particularly in heavy tasks such as multi-label classification. This question fundamentally motivates the development of quasi-multimodal diagnostic system, one that enables deterministic feature parsing and inherently promotes a few-shot generalization paradigm.

\subsection{Quasi-multimodal diagnosis}
Quasi-multimodal diagnosis, which leverages complementary information from multiple data sources, has emerged as a prominent topic in medical imaging analysis. Li et al. proposed a 3D fully convolutional network to rationally fuse the complementary information in PET/CT for accurate tumor segmentation \cite{Li2020a}. Zhang et al. adopted the generative adversarial network learning scheme to learn cross-modal features and developed a novel deep neural network that utilized these features to enhance the performance of brain tumor segmentation \cite{Zhang2021}. This approach highlighted the potential of cross-modal learning in improving diagnostic outcomes. In retinal disease diagnosis, Li et al. explored the potential of using paired color fundus photography (CFP) and fluorescein fundus angiography (FFA) images to improve fundus disease classification, aiming to exploit FFA’s ability to detect pathological changes such as non-perfusion areas and vascular leakage \cite{Li2020b}. Additionally, Hervella et al. employed unlabeled multimodal data for pretraining, including CFP and FFA images, to improve the segmentation of the optic disc and cup, and early diagnosis of related diseases, e.g., glaucoma, and the explainability \cite{Hervella2020}.

In most quasi-multimodal models, the number of incorporated modalities is limited. Intuitively, the more modalities involved, the more diagnostic accuracy could be expected. This conclusion is demonstrated in our results (see section 4.3). However, as the number of modalities increase, issues like information redundancy and mutual exclusion need to be seriously concerned. In our method, the multimodal features are adaptively calibrated depending on the disease being interrogated.

\section{Methodology}

\subsection{Overview}

\begin{figure}[H]
    \centering
     \includegraphics[width=1.0\linewidth]{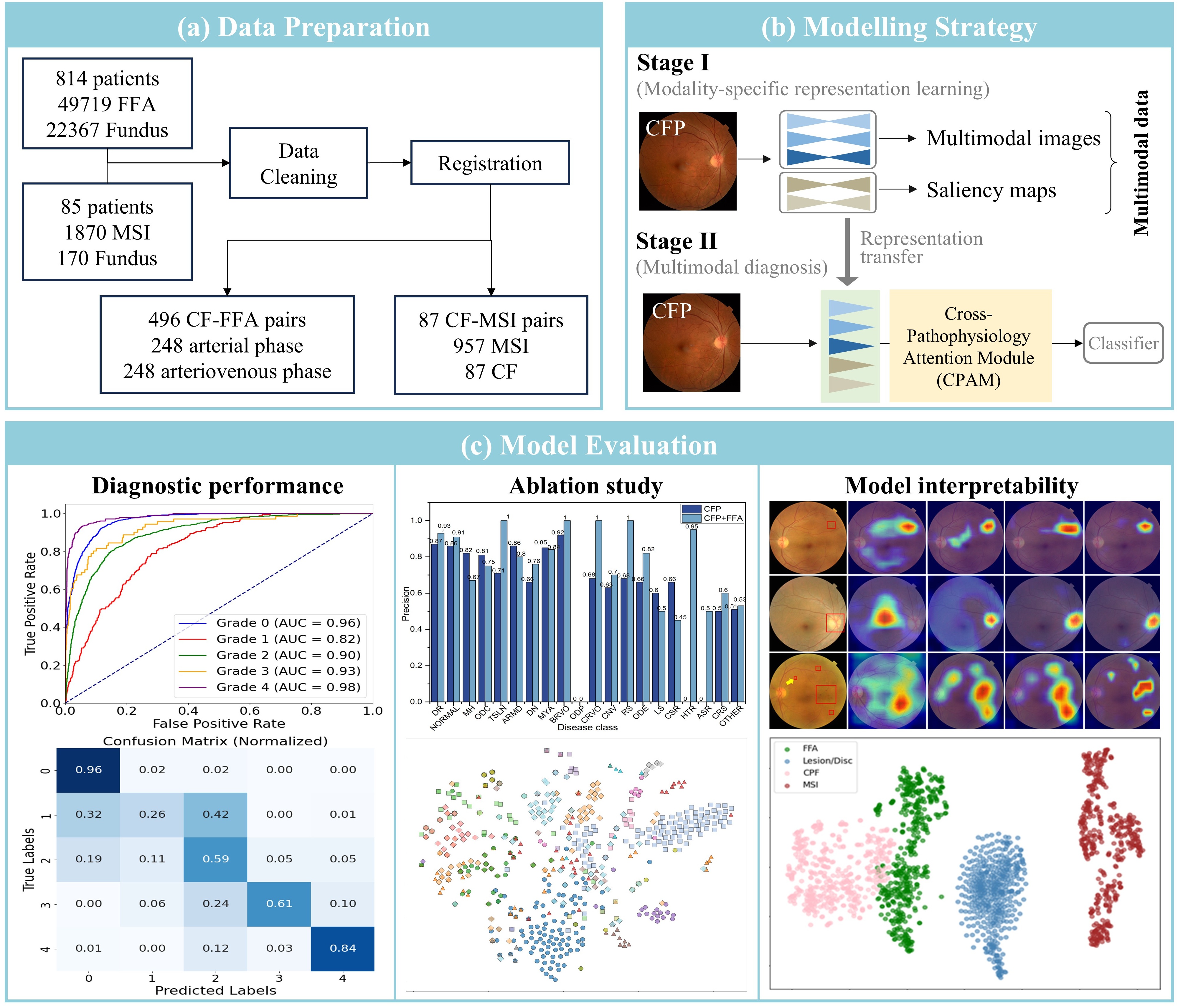} 
    \caption{Overview of the study: (a) data collection and preprocessing, (b) two-stage implementation of the modelling strategy, and (c) evaluation and interpretation.}
    \label{fig:overview}
\end{figure}

Fig.\ref{fig:overview} presents an overview of the study design, including data preparation, modelling strategy, and model evaluation. In the phase of data preparation, a portion of the dataset was collected from Tianjin Eye Hospital, comprising FFA images from 814 patients and MSI images from 85 patients. The remaining portion was obtained from three public sources, i.e., DDR \cite{Li2019b}, REFUGE \cite{Orlando2020} and MuReD \cite{Rodriguez2022}. The proposed method is implemented in two stages. In Stage I, multimodal data that involves multimodal images and saliency maps is learned from CFP images alone. The learned representations are then transferred to Stage II to formulate a multimodal diagnostic pattern. Details in stages I and II are illustrated in Sections 3.2 and 3.3, respectively. The effectiveness of our method was quantitatively validated for both multi-label classification and DR grading, with the significance and contribution of each network component and calibration node evaluated and interpreted.

\begin{figure}[H]
    \centering
     \includegraphics[width=1.0\linewidth]{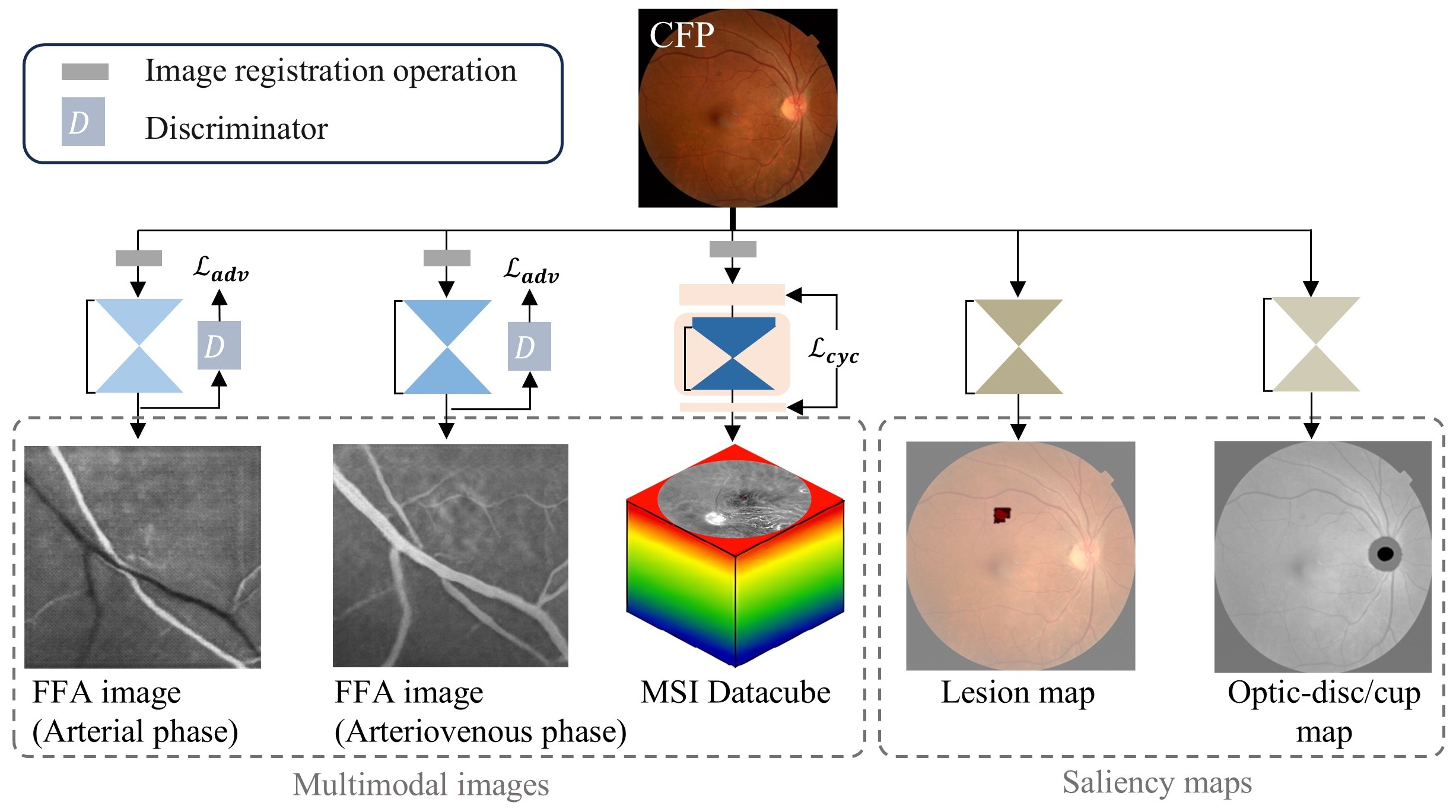} 
    \caption{Illustration of modality-specific representation learning. Parallel models, uniformly backboned on an encoder-decoder structure, are utilized to map CFP to multiple modalities across the pathophysiological spectrum. Representative losses used to regularize multimodal images synthesis are indicated, including adversarial loss ($\mathcal{L}_{adv}$) and cycle-consistency loss ($\mathcal{L}_{cyc})$}.
    \label{fig:2}
\end{figure}

\subsection{Modality-specific representation learning in Stage I}
As shown in Fig.\ref{fig:2}, parallel models uniformly based on an encoder-decoder backbone are employed to learn mapping from CFP to MSI, FFA in both arterial and arteriovenous phases, and saliency maps that segment latent lesion and optic-disc/cup regions in CFP. The synthesis process for each modality is described thoroughly below.

\textbf{FFA image synthesization:} Arteries and veins in CFP images exhibit distinct characteristics in terms of color, reflectance, and physical size, which provide a physiological basis for synthesizing FFA images in both arterial and arteriovenous phases. Given the limited area of vascular structure relative to the entire fundus, supervised learning with pixelwise corresponds is performed after careful alignment in between FFA and CFP image pairs. To accommodate any residual misalignment, an image-to-image translation network (pix2pix) is employed as the backbone. Further implementation details are provided in Appendix A, section 1.

The dataset was sourced from Tianjin Eye Hospital, comprising 49,719 CF images and 22,367 FFA images from 814 patients. Images not captured during the arterial phase or with severe blur were excluded. After screening, 248 matched pairs of arterial and arteriovenous phase FFA images were collected. All patient data were anonymized and de-identified in accordance with the Declaration of Helsinki.

\textbf{MSI image synthesization:}Biological tissues exhibit relatively low absorption and scattering within the near-infrared window, allowing for deeper penetration and higher contrast. As such, MSI datacube was synthesized over an extended spectral range, covering from visible band up to near-infrared. The feasibility of mapping RGB to near-infrared data has been demonstrated previously \cite{Wang2017a}\cite{Xiong2023}. 
   
   MSI synthesization is achieved using a backbone based on DRCR-Net \cite{Li2022}, with the network architecture described in Appendix A, Section 2. In practice, the spectral response function (SRF) of the imaging sensor in CFP devices varies significantly and is rarely measured precisely. To circumvent this limitation and enhance the generalizability, SRF randomization is introduced during training data generation. Specifically, randomized SRFs are generated by perturbing a baseline SRF measurement from the CFP device used in this study, as defined in Eq.(A.2). To further improve training stability and convergence, a cycle-consistency loss ($\mathcal{L}_{{cyc}}$) is imposed between the input CFP image and the CFP image reconstructed from the synthesized MSI datacube using the SRF associated with the input CFP, with the reconstruction procedure defined in Eq.(A.5).
   
   The dataset was obtained from Tianjin Eye Hospital, including both left (OS) and right (OD) eyes from 85 patients, yielding a total of 170 datacube sets. All examinations were performed using the same MSI instrument (RHA2020, Annidis Corporation). After removing those with strong reflections, highly uneven illumination, or large misalignments, 87 datacube sets were retained. Each datacube involved 11 images captured at wavelengths of 550, 580, 590, 620, 660, 685, 740, 760, 780, 810, and 850 nm. Images beyond 760nm were discarded given the large gap from the visible band. All images were roughly aligned to the one at 550 nm to facilitate convergence and robust against artifacts and noise.

\textbf{Saliency map synthesization:}Saliency maps here refer to those highlighting latent lesions and optic- disc/cup, which are critical for early detection, diagnosis, and monitoring of disease progression in ophthalmologic practice \cite{Wang2021b}. The saliency maps were represented in the form of image segmentation. For lesion extraction, the model was trained on the DDR dataset \cite{Li2019b} to identify key lesion types, including hemorrhage (HE), hard exudate (EX), soft exudate (SE), and microaneurysm (MA). For extraction of optic- disc and cup, training was performed on the REFUGE dataset \cite{Orlando2020}. Two separate models, both based on a U-Net backbone, were independently trained for lesion and disc/cup segmentation (See Appendix A, section 3).

\subsection{Multimodal diagnosis in Stage II}

In Stage II, the learned representations are transferred as image features for multimodal diagnosis. The encoders from the parallel models are isolated with potential shortcuts removed, and served as feature extractors of pathophysiological information. The extracted representative features are fused through adaptive and hierarchical calibration, tailored to the downstream diagnostic task. To further align the image synthesis process with diagnostic objectives, the encoders responsible for generating MSI and FFA images (in both arterial and arteriovenous phases) are finetuned jointly with the training of the fusion pipeline, under the combined supervision of a task-specific loss and fidelity regularizations.

\begin{figure}[H]
    \centering
     \includegraphics[width=1.0\linewidth]{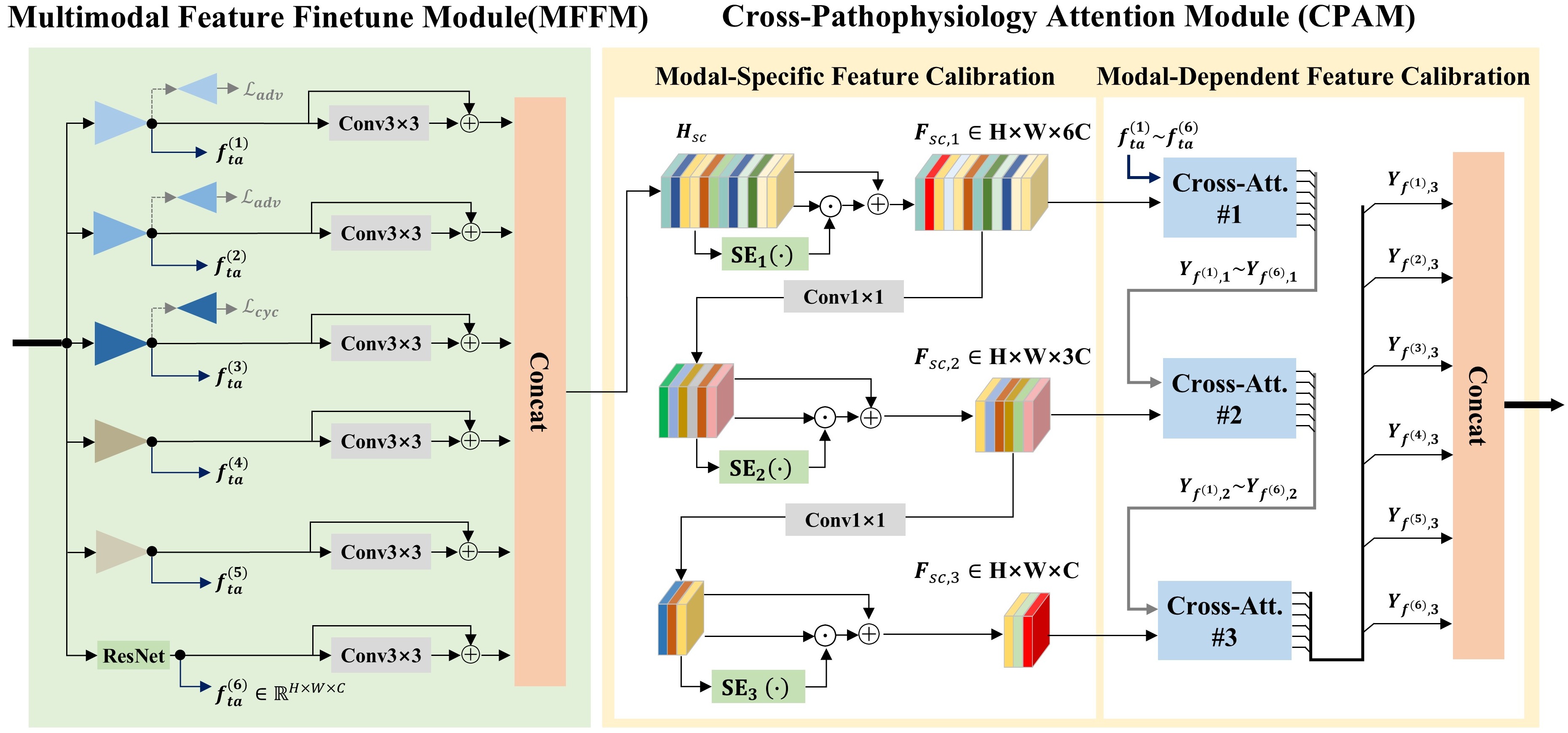} 
    \caption{Architecture of the multimodal diagnostic system. The system consists of a Multimodal Feature Finetuning Module (MFFM) followed by a Cross-Pathophysiology Attention Module (CPAM). Multimodal features generated by the encoders are first refined by the MFFM, which jointly finetunes these encoders along with the overall diagnostic pipeline to align image synthesis with diagnostic objectives. The refined features are then further calibrated by the CPAM, which employs a hierarchically cascaded attention mechanism to enable fine-grained, multi-scale alignment of pathophysiological features across modalities.}.
    \label{fig:3}
\end{figure}

\textbf{Multimodal Feature Finetune Module (MFFM):}To improve the sensitivity and representativity for the diagnosis purposes, the encoders associated with FFA and MSI synthesis are finetuned in a task-driven manner. To alleviate the need for large-scale paired training data, only a subset of Stage I losses that support unsupervised or self-supervised learning are retained during finetuning. Specifically, the decoders pretrained in Stage I are frozen and used to reconstruct FFA or MSI images, which are further constrained by selected losses, such as adversarial loss ($\mathcal{L}_{\text{adv}}$) and cycle-consistency loss ($\mathcal{L}_{\text{cyc}}$), as indicated in Fig.\ref{fig:3}. These losses help preserve visual fidelity while adapting the encoders for diagnostic tasks. The encoders for saliency map synthesis remain frozen to preserve their original representational integrity. To mitigate latent overfitting or optimization challenges, each encoder for the m-th modality is attached to a residual learning block $R_m(\cdot)$. Additionally, an independent feature extractor based on ResNet-18 is introduced to preserve the raw and modality-specific semantic information within CFP images. The output of this node is the concatenation of all extracted features $f_{\text{ta}}^{(m)} (\in \mathbb{R}^{H \times W \times C})$ 

\begin{equation}
\boldsymbol{H}_{sc} := \text{ReLU}\!\left(\left[R_1\!\left(\boldsymbol{f}_{ta}^{(1)}\right); R_2\!\left(\boldsymbol{f}_{ta}^{(2)}\right); \dots; R_6\!\left(\boldsymbol{f}_{ta}^{(6)}\right)\right]\right) \in \mathbb{R}^{H \times W \times 6C}.
\end{equation}

All the feature extractions operate in a plug-and-play manner, enabling concurrent integration of relevant information for effective pathophysiological attention. To fully exploit consistent and complementary information within and across the modalities, a Cross-Pathophysiology Attention Module (CPAM) is proposed to tackle with the representative features, which consists of two sub-modules, i.e., Modal-Specific Feature Calibration (MSFC) and Modal-Dependent Feature Calibration (MDFC).

\textbf{Modal-Specific Feature Calibration (MSFC):}Each modality carries unique characteristics linked to one or more diseases, which may not be fully captured in the high-level features ($f_{\text{ta}}^{(m)}$). Therefore, MSFC is designed to capture semantic information unique to each modality within the feature space. As illustrated in Fig.\ref{fig:3}, the finetuned feature vector ($\boldsymbol{H}_{sc}$) is fed into MSFC, which comprises three hierarchically stacked squeeze-and-excitation (SE) pipelines. These pipelines perform progressive, multi-scale calibration, allowing adaptive recalibration of feature responses at different levels of abstraction. By leveraging both global and contextual information, MSFC selectively emphasizes disease-relevant features while suppressing redundant or irrelevant responses. Each SE module is defined as

\begin{equation}
  \mathrm{SE}_k(\cdot) := \mathrm{Sigmoid}\!\left[\boldsymbol{\Theta}_{2,k} \, \mathrm{ReLU}\!\left(\boldsymbol{\Theta}_{1,k} P(\cdot)\right)\right], \quad k \in \{1,2,3\},
\end{equation}

where $P(\cdot)$ denotes global average pooling,$\boldsymbol{\Theta}_{1,k} \in \mathbb{R}^{(C/r) \times 6C} \quad \text{and} \quad \boldsymbol{\Theta}_{2,k} \in \mathbb{R}^{6C \times (C/r)}$are the weights of two fully connected layers, and r is the reduction ratio, empirically set to 16. The output from the k-th (k>1) SE pipeline is

\begin{equation}
  \boldsymbol{F}_{sc,k} := conv\!\left(\boldsymbol{F}_{sc,k-1}\right) \odot \mathrm{SE}_k\!\left[conv\!\left(\boldsymbol{F}_{sc,k-1}\right)\right] + conv\!\left(\boldsymbol{F}_{sc,k-1}\right)
\end{equation}

where $conv(\cdot)$ denotes a 1×1 convolution; and $\odot$ represents Hadamard product.

\begin{figure}[H]
    \centering
     \includegraphics[width=1.0\linewidth]{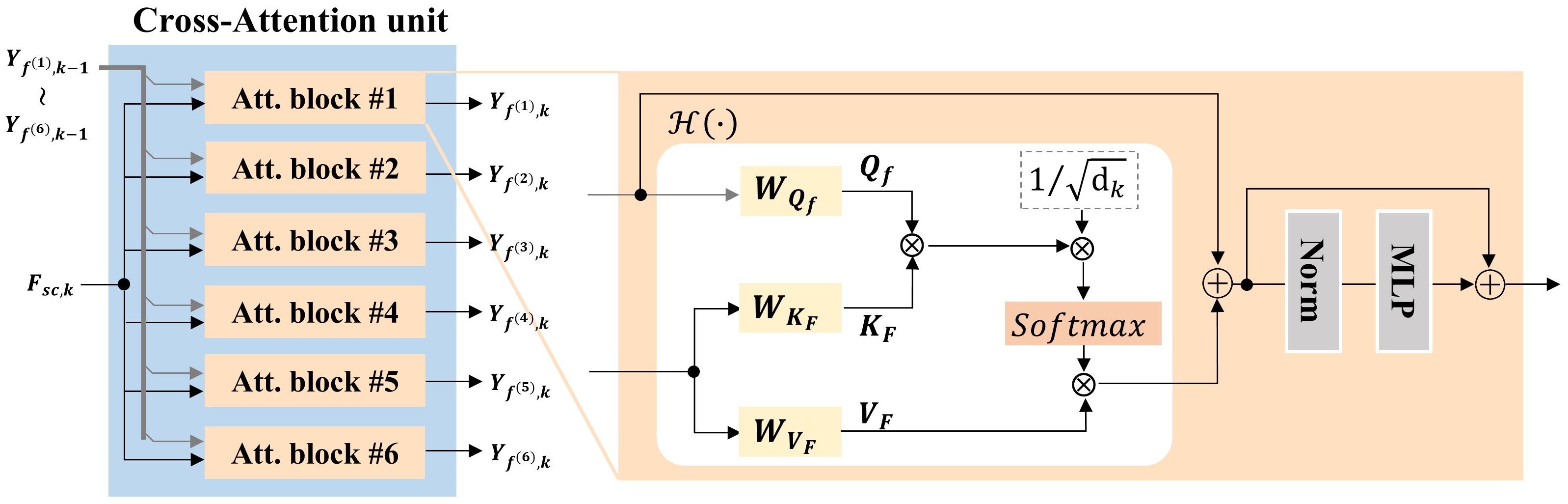} 
    \caption{Architecture of the k-th cross-attention unit in the phase of cross-modal feature calibration. Norm represents layer normalization,$\oplus$ and $\otimes$ represents matrix addition and matrix multiplication, respectively.}
    \label{fig:4}
\end{figure}

\textbf{Modal-Dependent Feature Calibration (MDFC):}For a given disease, multiple pathophysiological characteristics could be simultaneously reflected across these modalities, albeit with varying significance. To effectively capture such heterogeneous yet complementary cues, MDFC refines MSFC outputs $\boldsymbol{F}_{sc,k}$ through recursive, cross-modal interactions at multiple scales k(=1,2,3). Specifically, the MSFC output at each scale is fed into a scale-aligned cross-attention unit. As illustrated in Fig.\ref{fig:4}, each unit integrates six stacked attention blocks, enabling deep interaction and iterative refinement of inter-modal features. This recursive attention mechanism promotes more discriminative feature representations \cite{Praveen2023}\cite{Praveen2024}. Within each attention block, every individual "token"$\boldsymbol{Y}_{f^{(m)},k-1}$ interacts with all other "tokens" $\boldsymbol{F}_{sc,k}$ one at a time, capturing long-range dependencies.
Specifically, $\boldsymbol{F}_{sc,k}$ and $\boldsymbol{Y}_{f^{(m)},k-1}$ are reshaped to $\phi \times HW$ ($\phi \in \{6C,3C,C\}$), and processed through an attention head, defined as

\begin{equation}
  \mathcal{H}\!\left(\boldsymbol{F}_{sc,k}, \boldsymbol{Y}_{f^{(m)},k-1}\right) := \mathrm{softmax}\!\left(\frac{\boldsymbol{Q}_Y \boldsymbol{K}_F}{\sqrt{\phi}}\right) \boldsymbol{V}_F,
\end{equation}

where $\boldsymbol{Q}_Y$, $\boldsymbol{K}_F,$and $\boldsymbol{V}_F$ are the query, key and value matrices, respectively, given by
\[
\begin{cases}
\boldsymbol{Q}_Y := \boldsymbol{Y}_{f^{(m)},k-1} \boldsymbol{W}_{Q_Y} \\
\boldsymbol{K}_F := \boldsymbol{F}_{sc,k} \boldsymbol{W}_{K_F}  ,  \\
\boldsymbol{V}_F := \boldsymbol{F}_{sc,k} \boldsymbol{W}_{V_F}
\end{cases}
\tag{5}
\]

with $\boldsymbol{W}_{Q_Y}$, $\boldsymbol{W}_{K_F},$and $\boldsymbol{W}_{V_F}$ being the learnable projection matrices. The inputs to the first cross-attention unit are $\boldsymbol{F}_{sc,k}$ and $f_{\text{ta}}^{(m)}$.The attention output passes through a normalization layer and a Multi-Layer Perceptron (MLP) with residual connections to enhance nonlinear representation and stabilize training. Finally, outputs from all modalities($\boldsymbol{Y}_{f^{(m)},3}$)are concatenated to form $\boldsymbol{Y}$, which is fed to a classifier head for ultimate prediction.

The overall learning objective in Stage II is formulated as

\begin{equation}
  \mathcal{L}_{total} = \mathcal{L}_{task} + \lambda_1 \mathcal{L}_{cyc} + \lambda_2 \mathcal{L}_{adv}
  \tag{6},
\end{equation}

where $\mathcal{L}_{\text{adv}}$ and $\mathcal{L}_{\text{cyc}}$ are the fidelity-related regularization terms that preserve the perceptual realism of the synthesized FFA and MSI images, and $\mathcal{L}_{\text{task}}$ is the task-specific loss (defined in the experimental settings). The weighting factors $\lambda_1$ and $\lambda_2$ were both empirically set to 0.2.

\subsection{Datasets}
\begin{figure}[H]
    \centering
     \includegraphics[width=1.0\linewidth]{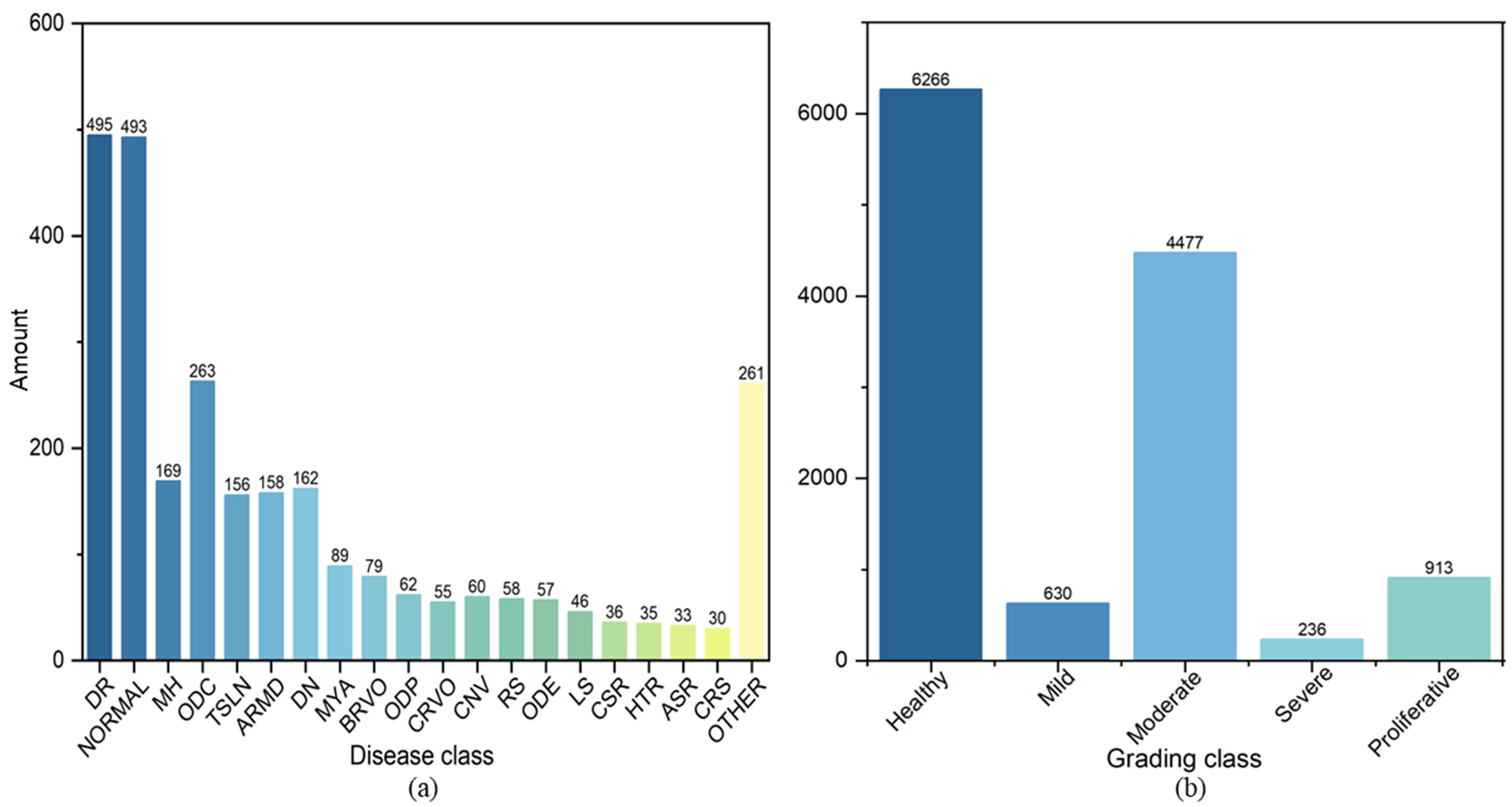} 
    \caption{Histograms of disease and DR grade in the datasets of (a) MuReD and (b) DDR, respectively.}
    \label{fig:5}
\end{figure}

Two publicly available datasets, i.e., MuReD \cite{Rodriguez2022}) and DDR \cite{Li2019b}, were employed for verification purposes, specifically for multi-label classification and DR grading, respectively. The MuReD dataset integrates three sub-datasets, i.e., ARIA, STARE, and RFMiD, comprising a total of 2,208 samples annotated with 20 distinct disease labels. Among these samples, 1,764 are reserved for training and validation, with the remaining for testing. The DDR dataset comprises 12,522 fundus images, divided into training (6,260 images), validation (2,503 images), and testing (3,759 images) sets. These retinal images were categorized into five grades based on the severity of diabetic retinopathy (DR): healthy, mild, moderate, severe, and proliferative. Fig.\ref{fig:5} shows the class distribution in the MuReD and DDR datasets.

\subsection{Implementation Details}
Training in both Stages I and II follow the same optimization strategy, i.e., the Adam optimizer (Kingma and Ba, 2014) with parameters $\beta_1$= 0.5, $\beta_2$ = 0.999. Detailed training procedures for each modality-specific representation in Stage I are provided in Sections 1–3 of the Appendix A. In Stage II, data were preprocessed using standard augmentation techniques such as horizontal flipping, vertical flipping, and random rotation to mitigate overfitting. The learning rate was initialized at 0.0001 and dynamically adjusted using the cosine annealing. The batch size was configured to 16. Our framework was implemented in PyTorch, and all experiments were conducted on a single NVIDIA RTX 4090 GPU.

\section{Experimental results}
\subsection{Multi-label classification}
The proposed model was trained on the MuReD dataset using asymmetric loss ($\mathcal{L}_{\text{task}}$) \cite{Ridnik2021} with a 5-fold cross-validation for fair comparison. Performance was evaluated against state-of-the-art (SOTA) approaches in terms of precision, F1-score, AUC, and mAP. All SOTA models were initialized using the ImageNet-1k pretrained weights and finetuned for 200 epochs, except RETFound, which used specifically pretrained weights and was finetuned for 20 epochs \cite{Zhou2023}. All methods shared the same experimental settings. Each metric was computed as the average over 20 diseases, including diabetic retinopathy(DR), normal retina(NORMAL), media haze(MH), optic disc cupping(ODC), tessellation(TSLN), age-related macular degeneration(ARMD), drusen(DN), myopia(MYA), branch retinal vein occlusion(BRVO), optic disc pallor(ODP), central retinal vein occlusion(CRVO), choroidal neovascularization(CNV), retinitis(RS), optic disc edema(ODE), laser scars(LS), central serous retinopathy(CSR), hypertensive retinopathy(HTR), arteriosclerotic retinopathy(ASR), chorioretinitis(CRS), and other diseases(OTHER). As tabulated in Table \ref{tab:multi_label_comparison}, our method significantly outperforms all competitors, achieving improvements of 16.57\%,\ 4.11\%,\ 1.92\%,\ and 2.02\%\ in precision, F1-score, AUC, and mAP, respectively.

\begin{table}[ht]
\centering
\caption{Comparisons with state-of-the-art approaches for multi-label classification}
\label{tab:multi_label_comparison}
\begin{tabular*}{0.6\linewidth}{@{\extracolsep{\fill}} ccccc @{}}
\toprule[1pt]
\textbf{Approach} & \textbf{Precision} & \textbf{F1} & \textbf{AUC} & \textbf{mAP} \\
\midrule
InceptionV3\,\cite{Szegedy2016} & 0.661 & 0.489 & 0.881 & 0.435 \\
ResNet-18\,\cite{He2016} & 0.614 & 0.576 & 0.894 & 0.452 \\
DenseNet121\,\cite{Huang2017} & 0.611 & 0.583 & 0.929 & 0.545 \\
Tresnet-r\,\cite{Ridnik2021} & 0.597 & 0.605 & 0.893 & 0.557 \\
ML-GCN\,\cite{Chen2019} & 0.697 & 0.586 & 0.912 & 0.645 \\
RETFound\,\cite{Zhou2023} & 0.720 & 0.603 & 0.927 & 0.652 \\
C-Tran\,\cite{Rodriguez2022} & 0.761 & 0.656 & 0.945 & 0.740 \\
MedMamba\,\cite{Yue2024} & 0.803 & 0.548 & 0.940 & 0.633 \\
LAGNet\,\cite{Liang2024} & 0.786 & 0.643 & 0.936 & 0.644 \\
IRECTe\,\cite{Singh2025} & 0.718 & 0.656 & 0.935 & 0.742 \\
Ours & \textbf{0.837} & \textbf{0.683} & \textbf{0.953} & \textbf{0.757} \\
\bottomrule[1pt]
\end{tabular*}
\end{table}

To offer a deeper insight into the model performance, Table \ref{tab:per_class_precision} reports the precision scores for each specific disease as mean ± standard deviation. Overall, our method demonstrates statistically significant improvements over SOTA ones across 11 disease classes (p<0.05). Specifically, for diseases characterized by vascular changes such as BRVO, CRVO and CNV, the precision improves by 8.07\%,\ 1.69\%\ and 35.41\%,\ respectively, compared to IECTe \cite{Singh2025}. This probably is attributed to the integration of FFA-derived features, which emphasize the fundus vascular structures. For diseases involving characteristic choroidal pathology, such as TSLN, DN and MYA, the precision increases by 18.42\%,\ 45.40\%\ and 29.41\%,\ respectively. While the underlying reasons for these improvements could be multifaceted, the contributions from MSI particularly in bands beyond the visible spectrum could be substantial due to the relatively high penetration depth, reaching up to the choroid layer. Benefiting from the incorporation of saliency map, the model also shows marked improvements in detecting conditions directly related to the optic disc and cup, such as ODC and ODE. This is because the segmentation-based saliency features are inherently sensitive to structural and boundary cues, thereby enhancing the recognition of diseases with prominent shape deformations. In contrast, their contribution to ODP is limited, as ODP primarily manifests with subtle color and brightness variations rather than structural deformation. Although the diagnostic performance for ASR remains suboptimal (precision: 0.250), other listed methods almost fail to detect it (precision < 0.001). This may be due to the small size of training set (33 ASR cases), which exacerbates diagnostic uncertainty given the phenotypic similarity between ASR and other vascular diseases. In contrast, CSR yields better results despite a similarly limited size of samples (36 cases), probably because its distinct pathological signatures reduce confusion with other retinal diseases. 

\begin{table}[ht]
\centering
\caption{Analysis for per-class classification in terms of precision score}
\resizebox{\textwidth}{!}{
\begin{tabular}{lcccccccc}
\toprule[1.2pt]
Pathology & Tresnet-L & ML-GCN & RETFound & C-Tran & MedMamba & LAGNet & IECTe & \textbf{Ours} \\
\midrule
DR & 0.787\textpm0.041 & 0.785\textpm0.039 & 0.786\textpm0.087 & 0.869\textpm0.061 & 0.865\textpm0.076 & 0.870\textpm0.041 & 0.814\textpm0.075 & \textbf{0.891\textpm0.018} \\
NORMAL & 0.556\textpm0.055 & 0.748\textpm0.045 & 0.736\textpm0.040 & 0.723\textpm0.051 & 0.820\textpm0.043 & 0.758\textpm0.061 & 0.746\textpm0.039 & \textbf{0.862\textpm0.013*} \\
MH & 0.545\textpm0.178 & 0.785\textpm0.052 & 0.768\textpm0.050 & 0.722\textpm0.110 & 0.709\textpm0.135 & 0.603\textpm0.097 & 0.765\textpm0.056 & \textbf{0.865\textpm0.050*} \\
ODC & 0.353\textpm0.183 & 0.614\textpm0.076 & 0.668\textpm0.088 & 0.544\textpm0.190 & 0.803\textpm0.052 & 0.655\textpm0.177 & 0.648\textpm0.088 & \textbf{0.832\textpm0.025*} \\
TSLN & 0.500\textpm0.047 & 0.747\textpm0.120 & 0.711\textpm0.103 & 0.754\textpm0.100 & 0.633\textpm0.121 & 0.805\textpm0.040 & 0.722\textpm0.112 & \textbf{0.855\textpm0.033*} \\
ARMD & 0.419\textpm0.185 & 0.617\textpm0.218 & 0.680\textpm0.185 & 0.694\textpm0.166 & \textbf{0.800\textpm0.244} & 0.787\textpm0.065 & 0.656\textpm0.200 & 0.717\textpm0.084 \\
DN & 0.218\textpm0.138 & 0.468\textpm0.099 & 0.546\textpm0.129 & 0.631\textpm0.129 & 0.509\textpm0.183 & 0.708\textpm0.030 & 0.511\textpm0.103 & \textbf{0.743\textpm0.063*} \\
MYA & 0.500\textpm0.074 & 0.707\textpm0.010 & 0.738\textpm0.054 & 0.841\textpm0.092 & 0.856\textpm0.034 & 0.874\textpm0.048 & 0.741\textpm0.060 & \textbf{0.959\textpm0.037*} \\
BRVO & 0.181\textpm0.090 & 0.873\textpm0.051 & 0.803\textpm0.098 & 0.650\textpm0.061 & 0.785\textpm0.163 & 0.912\textpm0.047 & 0.879\textpm0.080 & \textbf{0.950\textpm0.048*} \\
ODP & 0.100\textpm0.184 & \textbf{0.583\textpm0.117} & 0.267\textpm0.223 & 0.146\textpm0.180 & 0.250\textpm0.100 & 0.346\textpm0.184 & 0.387\textpm0.207 & 0.300\textpm0.217 \\
CRVO & 0.117\textpm0.138 & 0.833\textpm0.094 & 0.855\textpm0.120 & 0.850\textpm0.133 & 0.683\textpm0.166 & 0.843\textpm0.148 & 0.887\textpm0.113 & \textbf{0.902\textpm0.094*} \\
CNV & 0.381\textpm0.060 & 0.655\textpm0.122 & 0.663\textpm0.106 & 0.633\textpm0.071 & 0.541\textpm0.135 & 0.800\textpm0.077 & 0.641\textpm0.109 & \textbf{0.868\textpm0.039*} \\
RS & 0.500\textpm0.048 & 0.428\textpm0.137 & 0.819\textpm0.132 & 0.719\textpm0.051 & 0.400\textpm0.089 & 0.810\textpm0.135 & 0.802\textpm0.142 & \textbf{0.833\textpm0.102*} \\
ODE & 0.466\textpm0.105 & 0.565\textpm0.104 & 0.639\textpm0.175 & 0.721\textpm0.172 & 0.600\textpm0.139 & 0.766\textpm0.088 & 0.674\textpm0.198 & \textbf{0.875\textpm0.088*} \\
LS & 0.250\textpm0.160 & 0.622\textpm0.207 & 0.633\textpm0.171 & 0.366\textpm0.171 & 0.300\textpm0.200 & 0.400\textpm0.089 & \textbf{0.691\textpm0.216} & 0.533\textpm0.074 \\
CSR & 0.230\textpm0.181 & 0.200\textpm0.100 & 0.633\textpm0.171 & \textbf{0.766\textpm0.200} & 0.311\textpm0.194 & 0.700\textpm0.200 & 0.666\textpm0.208 & 0.632\textpm0.060 \\
HTR & 0.157\textpm0.214 & 0.000\textpm0.000 & 0.200\textpm0.100 & 0.400\textpm0.274 & 0.000\textpm0.000 & 0.200\textpm0.029 & 0.250\textpm0.233 & \textbf{0.900\textpm0.223*} \\
ASR & 0.000\textpm0.000 & 0.000\textpm0.000 & 0.000\textpm0.000 & 0.000\textpm0.000 & 0.000\textpm0.000 & 0.000\textpm0.000 & 0.000\textpm0.000 & \textbf{0.250\textpm0.282*} \\
CRS & 0.250\textpm0.214 & 0.250\textpm0.204 & 0.500\textpm0.253 & 0.340\textpm0.277 & 0.366\textpm0.133 & 0.333\textpm0.221 & \textbf{0.562\textpm0.169} & 0.468\textpm0.074 \\
OTHER & 0.365\textpm0.108 & 0.580\textpm0.056 & 0.595\textpm0.050 & 0.487\textpm0.066 & 0.285\textpm0.157 & 0.521\textpm0.087 & 0.583\textpm0.049 & \textbf{0.651\textpm0.048*} \\
\bottomrule[1.2pt]
\multicolumn{9}{@{}l}{\large $^{*}p<0.05$ compared with all baseline methods (paired $t$-test).}
\end{tabular}
}
\vspace{1mm}
\label{tab:per_class_precision}
\end{table}

\begin{figure}[H]
    \centering
     \includegraphics[width=1.0\linewidth]{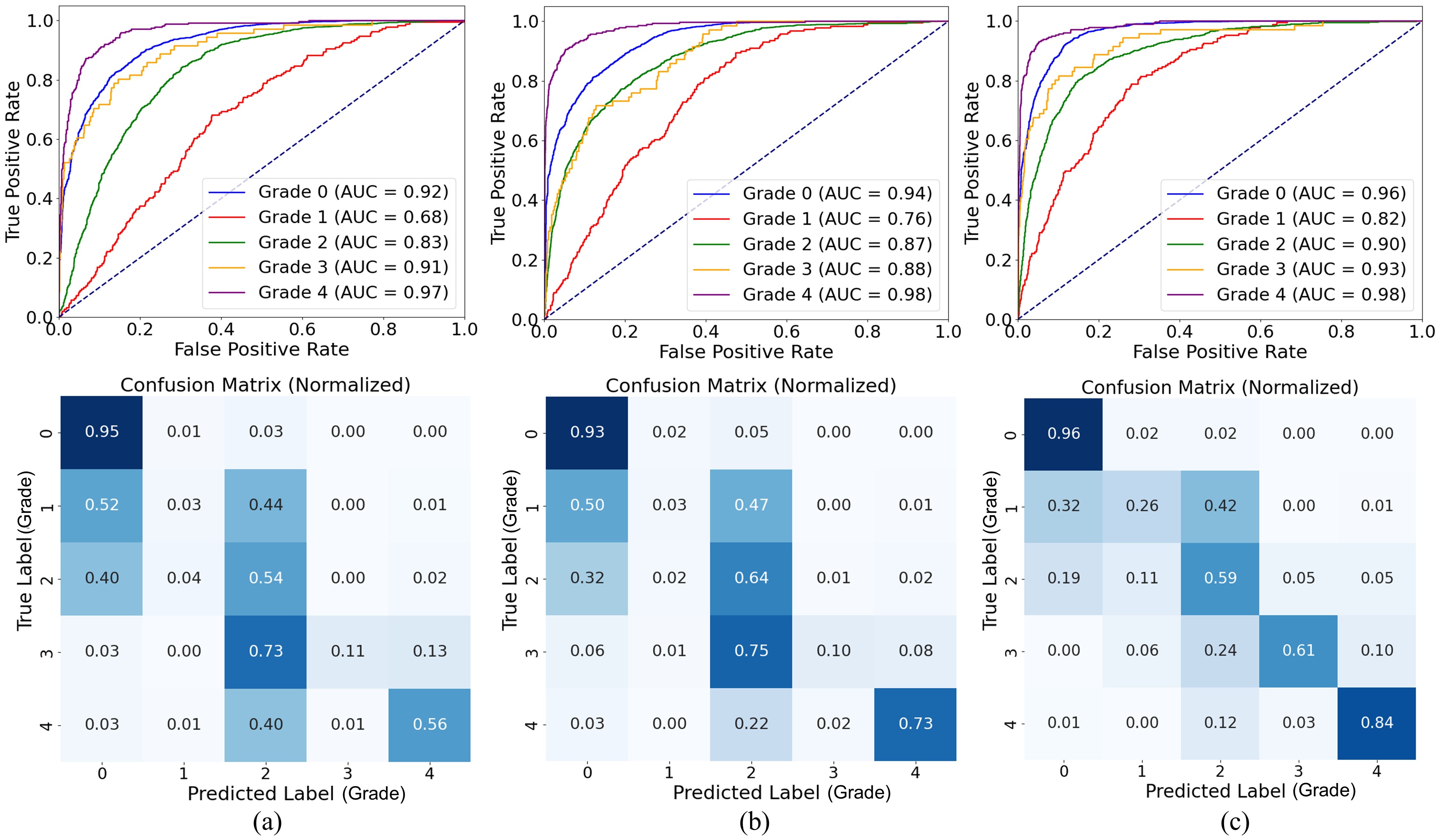} 
    \caption{Comparisons of ROC curves (top) and classification confusion matrices (bottom) among (a) DesNet121, (b) RETFound, and (c) ours, for DR grading.}
    \label{fig:6}
\end{figure}

Receiver operating characteristic (ROC) curves are presented in Fig.B.1 of Appendix B. Consistent with the precision results, our method achieves higher AUC values over SOTA ones across 14 diseases. Notably, for conditions characterized by vascular changes or specific choroidal pathology such as BRVO, CRVO, CNV, TSLN, and DN, our method outperforms IECTe, achieving AUC improvements of 1.12\%, 0.81\%, 0.81\%, 0.82\%, and 6.50\%, respectively.

\subsection{Diabetic retinopathy (DR) grading}
Our model was re-trained on the DDR dataset using categorical cross-entropy loss ($\mathcal{L}_{\text{task}}$). Although the DDR dataset was also employed in Stage I for saliency maps synthesis to highlight latent lesions, there was no overlap between the training and test datasets across the two Stages. The performance was compared with SOTA approaches, including M2CNN \cite{Zhou2018}, DeepMT-DR \cite{Wang2021b}, RETFound \cite{Zhou2023}, DRStageNet \cite{Men2023}, KA+KC-Net \cite{Tian2023}, MedMamba \cite{Yue2024}, ViT+CapsNet \cite{Oulhadj2024}), CRA-Net \cite{Zang2024}, and MuR-CAN (Madarapu et al., 2024), in terms of precision, accuracy, and the Cohen’s kappa coefficient (quadratic weighted). For methods with publicly available code, we reproduced results using the same initialization strategy as in Section 4.1 and identical experimental settings. Otherwise, results were taken directly from the original publications.

\begin{table}[ht]
\centering
\caption{Quantitative results for DR grading on the DDR dataset}
\label{tab:ddr_dr_grading}
\begin{tabular*}{0.7\linewidth}{@{\extracolsep{\fill}} ccccc @{}}
\toprule[1pt]
\textbf{Approach} & \textbf{Kappa} & \textbf{Precision} & \textbf{Accuracy} \\
\midrule
ResNet-18\,\cite{He2016} & 0.708 & 0.773 & 0.732 \\
Inception-v3\,\cite{Szegedy2016} & 0.719 & 0.789 & 0.744 \\
DenseNet-121\,\cite{Huang2017} & 0.689 & 0.792 & 0.740 \\
ViTCSRA\,\cite{Gu2023} & 0.732 & 0.757 & 0.769 \\
RETFound\,\cite{Zhou2023} & 0.776 & 0.741 & 0.776 \\
DRStageNet\,\cite{Men2023} & 0.779 & - & 0.779 \\
M²CNN\,\cite{Zhou2018} & 0.757 & 0.816 & 0.795 \\
KA+KC-Net\,\cite{Tian2023} & 0.826 & - & 0.821 \\
DeepMT-DR\,\cite{Wang2021b} & 0.802 & 0.831 & 0.836 \\
MedMamba\,\cite{Yue2024} & 0.820 & 0.801 & 0.801 \\
ViT+CapsNet\,\cite{Oulhadj2024} & - & 0.590 & 0.803 \\
CRA-Net\,\cite{Zang2024} & 0.840 & - & 0.831 \\
MuR-CAN\,\cite{Madarapu2024} & 0.836 & 0.831 & 0.837 \\
Ours & \textbf{0.861} & \textbf{0.835} & \textbf{0.842} \\
\bottomrule[1pt]
\end{tabular*}
\end{table}

As reported in Table \ref{tab:ddr_dr_grading}, our method consistently outperforms all competitors, surpassing the second-best method (MuR-CAN) by 0.48\%,\ 0.59\%,\ and 2.99\%\ in precision, accuracy, and Kappa, respectively. Receiver operating characteristic (ROC) curves and confusion matrices are represented in Fig.\ref{fig:6}. While all methods perform well in detecting grade 0 (no DR), DesNet121 and RETFound show noticeable drops for higher DR grades, e.g., grade 1 (mild DR) and grade 3 (severe DR). These degradations are evident in the confusion matrices, which reveal frequent misclassifications, e.g., grade 1 vs. grade 0 and grade 3 vs. grade 2. In contrast, our method achieves higher AUC values than RETFound by 2.12\%,\ 7.89\%,\ 3.44\%,\ and 5.68\%\ for grades 0–3, respectively, and correctly classifies more samples in nearly all grades, albeit with some intra-class confusion. These improvements could be interpreted through the synergistic integration of all input modalities. FFA provides detailed and dynamic visualization of retinal vasculature, known as the gold standard for DR diagnosis. Saliency maps focus attention on region-of-interest, effectively reducing false positives. MSI promotes tissue differentiation and identification of subtle pathological changes. The individual contributions of each modality are further examined in the ablation study. To assess generalizability, an external validation was conducted on the EyePACS dataset \cite{Wang2017b}\cite{Baliah2023}. A total of 1,000 images were randomly selected using stratified sampling to preserve the original DR grade distribution. As summarized in Table B.1, Appendix B, our method again outperforms the second-best competitor (MedMamba) by 5.53\%,\ 3.90\%,\ and 4.00\%\ in Kappa, precision, and accuracy, respectively.

\subsection{Ablation Study}
An ablation study was conducted to evaluate the individual contributions from each modality and each calibration node in our model, in the contexts of multi-label classification and DR grading.

\begin{table}[ht]
\centering
\caption{Ablation study on the importance of modalities}
\resizebox{\textwidth}{!}{
\begin{tabular}{|cccc|cccc|ccc|}
\hline
\multicolumn{4}{|c|}{\textbf{Modality}} & \multicolumn{4}{c|}{\textbf{Multi-label classification}} & \multicolumn{3}{c|}{\textbf{DR grading}} \\
\hline
CFP & FFA & MSI & Saliency Maps & Precision & F1 & AUC & mAP & Kappa & Precision & Accuracy \\
\hline
\checkmark & & & & 0.792 & 0.601 & 0.913 & 0.689 & 0.760 & 0.741 & 0.751 \\
\checkmark & \checkmark & & & 0.825 & 0.647 & 0.944 & 0.731 & 0.823 & 0.791 & 0.815 \\
\checkmark & & \checkmark & & 0.816 & 0.633 & 0.947 & 0.724 & 0.832 & 0.810 & 0.761 \\
\checkmark & \checkmark & \checkmark & & 0.827 & 0.656 & 0.950 & 0.733 & 0.842 & 0.822 & 0.834 \\
\checkmark & \checkmark & \checkmark & \checkmark & 0.837 & 0.683 & 0.953 & 0.757 & 0.861 & 0.835 & 0.842 \\
\hline
\end{tabular}
}
\label{tab:ablation_modality}
\end{table}

\begin{figure}[H]
    \centering
     \includegraphics[width=1.0\linewidth]{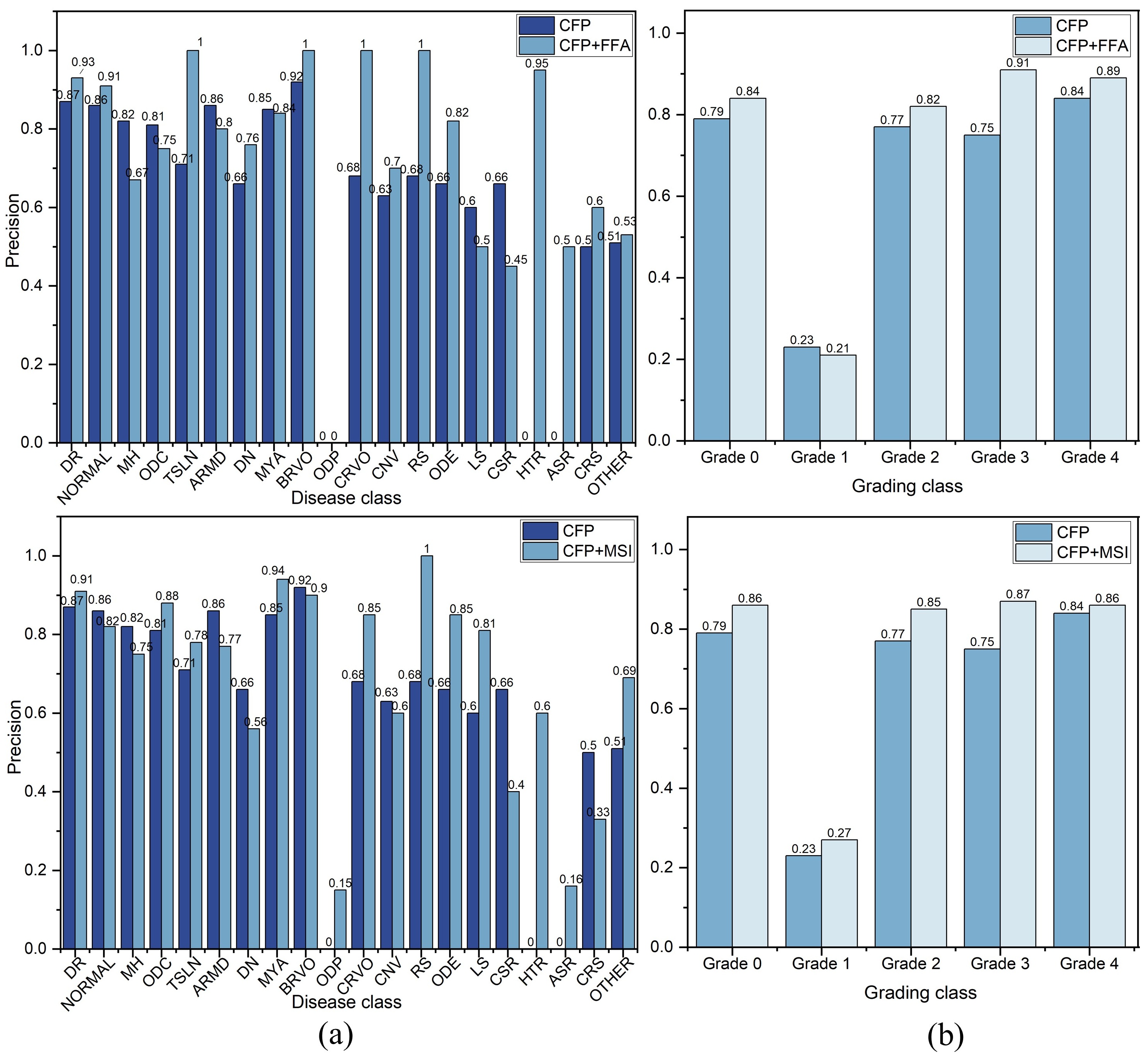} 
    \caption{Comparison of boosts from various modality combinations achieved for (a) multi-label classification and (b) DR grading. The top row compares CFP with CFP+FFA, while the bottom row compares CFP with CFP+MSI.}
    \label{fig:7}
\end{figure}

\textbf{Importance of Modality:}In the experiment, CFP was utilized as the baseline modality. The evaluation was initially performed on the MuReD dataset for the task of multi-label classification. As shown in Table \ref{tab:ablation_modality}, incorporating FFA into the baseline significantly enhances the performance, with mean increases of 4.16\%,\ 7.65\%,\ 3.39\%,\ and 6.09\%\ in precision, F1-score, AUC, and mAP, respectively. Specifically, improvements for vascular abnormality-related diseases such as BRVO and CRVO reach 8.69\%\ and 47.05\%,\ respectively, as illustrated in the top row of Fig.\ref{fig:7}(a). Further incorporation of MSI continues to improve model performance, although gains in metrics like precision and mAP remain modest. This outcome is reasonable, as most of the targeted diseases in this study involve circulatory disorders that are more effectively distinguished using modalities other than MSI. As anticipated, integrating both FFA and MSI into the baseline yields superior results over all the metrics, compared to any single-modality boost. Overall, the all-modality fusion provides the most comprehensive improvement, with mean increases of 5.68\%,\ 13.64\%,\ 4.38\%\, and 9.86\%\ in precision, F1-score, AUC, and mAP, respectively, compared to CFP alone. The average improvement across all the metrics reaches 8.39\%.\

The experiments were further conducted on the DDR dataset for DR grading. As observed from Table \ref{tab:ablation_modality}, a single boost from either FFA or MSI leads to a significant improvement in terms of Kappa, increased by 8.29\%\ and 9.47\%,\ respectively. On one hand, FFA provides detailed visualization of vascular structures and different types of vascular abnormalities, which is vital for early diagnosis and accurate assessment of disease severity. This is evidenced in the top row of Fig.\ref{fig:7}(b), where the integration of FFA boost leads to a 21.33\%\ improvement in precision for grade 3 diagnosis compared to using CFP alone. On the other hand, MSI improves the differentiation of retinal tissues and lesions that are often difficult to detect using a single wavelength or visible spectrum. Surprisingly, MSI boost can even outperform FFA in certain cases, e.g., an increase of 17.39\%\ in precision is observed for grade 1. Combining FFA and MSI provides complementary information from both vascular and tissue perspectives, which is critical for precise grading, especially in the early stages. This observation aligns with the results in Table \ref{tab:ablation_modality}, where accuracy improves by 11.05\%\ compared to the baseline. The overall improvement from all-modality boost reaches 12.69\%\ across all the metrics.

\begin{figure}[H]
    \centering
     \includegraphics[width=1.0\linewidth]{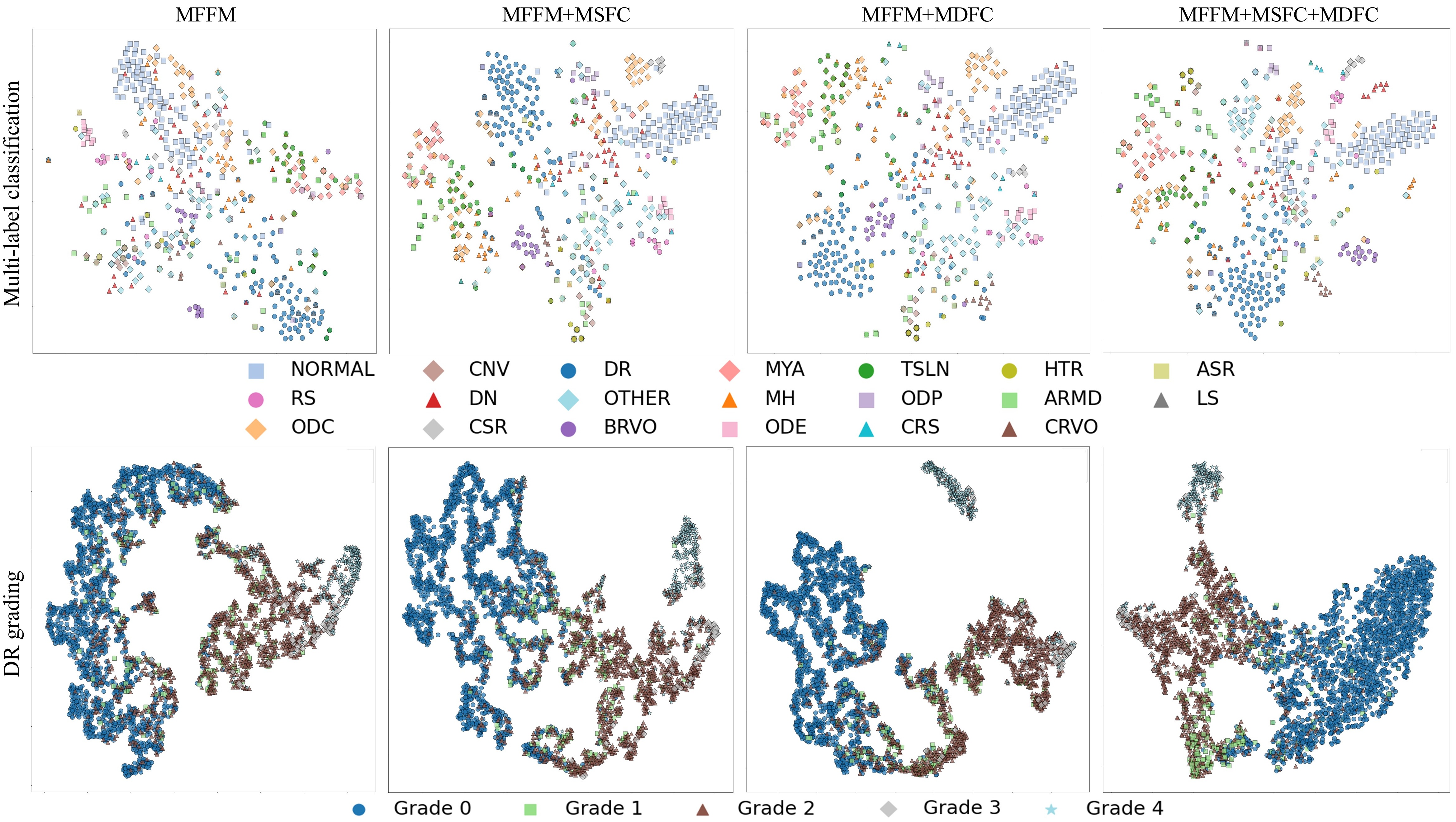} 
    \caption{Visualization of feature embedding casted for multi-label classification(top) and DR grading (bottom). The results for various calibration nodes are shown from left to right.}
    \label{fig:8}
\end{figure}

\textbf{Importance of Calibration Nodes:}As shown in Fig.\ref{fig:3}, multimodal features are sequentially calibrated via three nodes, i.e., junior calibration in MFFM, intermediate calibration in MSFC, and senior calibration in MDFC. In the baseline configuration, only MFFM is employed, with MSFC and MDFC removed. The significance of each calibration node is evaluated on the MuReD and DDR datasets, respectively. Quantitative results are summarized in Table \ref{tab:cpam_calibration_nodes}, with the associated feature embeddings visualized in Fig.\ref{fig:8}. As expected, the baseline calibration step yields the weakest performance, emphasizing the necessity of addressing challenges like information redundancy and mutual exclusion in multimodal diagnostic tasks. MFFM performs a straightforward cascading of transferred representations in feature space, assigning fixed weights to each modality. Therefore, cross-modality interactions could not be fully captured, which possibly deteriorates diagnosis on certain diseases, e.g., DN, MYA, and ODC. These diseases may benefit more significantly from particular imaging modalities. With the boost from either MSFC or MDFC, the model has a salient improvement over all the metrics, averaged at 3.56\%\ and 6.42\%\ on multi-label classification and DR grading tasks, respectively. These improvements are further visualized in Fig.\ref{fig:8}, where the addition of MSFC or MDFC results in clearer spatial clustering of disease and DR grade embeddings in latent space. Combining MSFC and MDFC yields further performance gains, e.g., precision in multi-label classification increases by 6.48\%.\ Besides, tighter clustering of the disease embeddings such as CSR, DN and BRVO is observed. These results demonstrate the capability to effectively calibrate cross-modal information and generate optimized semantic representations by dynamically emphasizing modality-specific diagnostic signatures.

\begin{table}[ht]
\centering
\caption{Quantitative evaluations of diverse combinations of the calibration nodes in CPAM}
\resizebox{\textwidth}{!}{
\begin{tabular}{|ccc|cccc|ccc|}
\hline
\multicolumn{3}{|c|}{\textbf{Node}} & \multicolumn{4}{c|}{\textbf{Multi-label classification}} & \multicolumn{3}{c|}{\textbf{DR grading}} \\
\hline
MFFM & MSFC & MDFC & Precision & F1 & AUC & mAP & Kappa & Precision & Accuracy \\
\hline
\checkmark & & & 0.786 & 0.617 & 0.914 & 0.676 & 0.758 & 0.774 & 0.791 \\
\checkmark & \checkmark & & 0.825 & 0.635 & 0.928 & 0.718 & 0.848 & 0.823 & 0.837 \\
\checkmark & & \checkmark & 0.807 & 0.636 & 0.933 & 0.710 & 0.854 & 0.814 & 0.826 \\
\checkmark & \checkmark & \checkmark & 0.837 & 0.683 & 0.953 & 0.757 & 0.861 & 0.835 & 0.842 \\
\hline
\end{tabular}
}
\label{tab:cpam_calibration_nodes}
\end{table}

\textbf{Importance of Fusion Strategies:}Several fusion strategies in Stage II are investigated by varying encoder status (frozen or unfrozen) and input type to CPAM (image-level or feature-level). For image-level fusion, synthesized images are fed directly into CPAM, whereas for feature-level fusion, CPAM receives inputs from MFFM. The proposed method adopts unfrozen encoders with feature-level inputs, corresponding to the last row in Table \ref{tab:fusion_strategies}. As presented in Table \ref{tab:fusion_strategies}, the worst performance occurs when encoders are frozen and image-level inputs are used. Switching to feature-level inputs yields substantial gains, i.e., average of 13.23\%\ and 12.31\%\ on the multi-label classification and DR grading tasks, respectively. Similar trends are observed when encoders are unfrozen, with improvements of 12.21\%\ and 11.46\%\ on the respective tasks. The superiority of feature-level fusion probably comes from reducing the gap between complex pixel-level synthesis and the relatively simpler nature of classification, as previously discussed. With inputs fixed at the feature level, further gains are achieved by unfreezing the encoders, including improvements of 2.95\%\ (precision), 3.17\%\ (F1), 1.18\%\ (AUC), and 3.55\%\ (mAP) on multi-label classification, alongside 1.58\%\ (precision), 1.32\%\ (accuracy), and 1.77\%\ (kappa) on DR grading. These results demonstrate that finetuning the encoders allows better alignment with diagnostic objectives, enabling more effective task-driven multimodal feature learning.

\begin{table}[ht]
\centering
\caption{Quantitative evaluations of diverse fusion strategies}
\resizebox{\textwidth}{!}{
\begin{tabular}{|cc|cc|cccc|ccc|}
\hline
\multicolumn{2}{|c|}{\textbf{Encoder status}} & \multicolumn{2}{c|}{\textbf{Multimodal sources}} & \multicolumn{4}{c|}{\textbf{Multi-label classification}} & \multicolumn{3}{c|}{\textbf{DR grading}} \\
\hline
Frozen & Unfrozen & Image-level & Feature-level & Precision & F1 & AUC & mAP & Kappa & Precision & Accuracy \\
\hline
\checkmark & & \checkmark & & 0.781 & 0.531 & 0.892 & 0.613 & 0.751 & 0.738 & 0.736 \\
\checkmark & & & \checkmark & 0.813 & 0.662 & 0.936 & 0.731 & 0.846 & 0.822 & 0.831 \\
& \checkmark & \checkmark & & 0.802 & 0.559 & 0.905 & 0.647 & 0.772 & 0.751 & 0.754 \\
& \checkmark & & \checkmark & 0.837 & 0.683 & 0.953 & 0.757 & 0.861 & 0.835 & 0.842 \\
\hline
\end{tabular}
}
\label{tab:fusion_strategies}
\end{table}

\subsection{Model interpretability}

\begin{figure}[H]
    \centering
     \includegraphics[width=1.0\linewidth]{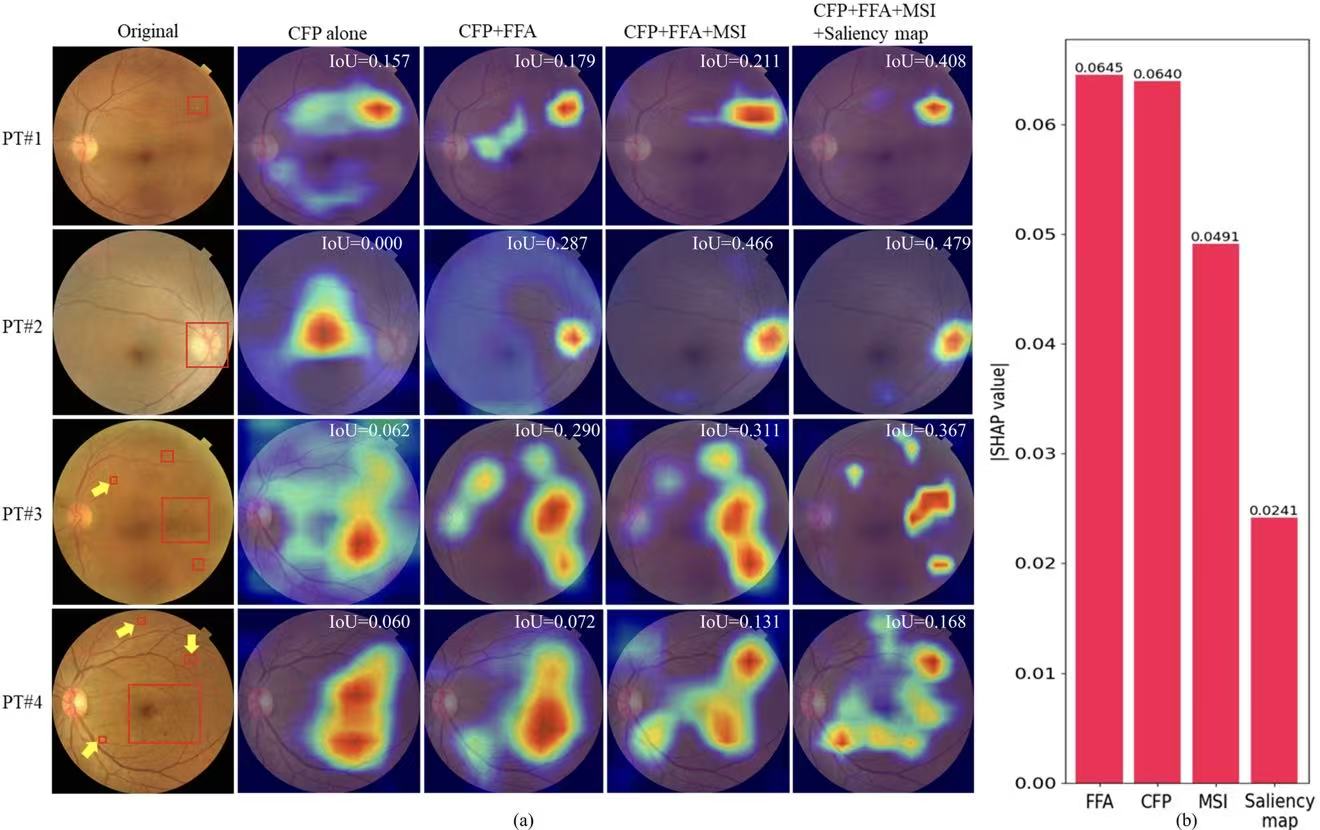} 
    \caption{Analysis of inter- and intra- modality (region-level) importance. (a) Grad-CAM visualizations on CFP images from representative patients with drusen (PT \#1), optic disc pallor (PT \#2), moderate NPDR (PT \#3), and severe NPDR (PT \#4). Key lesions and disease-related structures are annotated on the original fundus images. Highlighted regions indicate attention maps derived from different modality combinations, with IoU values indicating their overlap with the annotations. (b) SHAP values for each modality. The model was trained for multi-label classification.}
    \label{fig:9}
\end{figure}

To assess model interpretability and understand the contribution of each modality, attention heatmaps are visualized with Grad-CAM \cite{Selvaraju2017} and SHAP \cite{Lundberg2017}. As shown in Fig.\ref{fig:9}(a), the CFP images were collected from four representative patients diagnosed with drusen, optic disc pallor (ODP), moderate non-proliferative diabetic retinopathy (NPDR), and severe NPDR, numbered as PT \#1-4, respectively. Latent lesions in the original CFP images were annotated by three senior ophthalmologists. The overlaps between the annotated lesions and Grad-CAM attention maps were quantified in terms of the Intersection over Union (IoU) for each modality combination.

For PT \#1, identifying smaller drusen using CFP alone remains challenging, as attention maps tend to highlight several irrelevant regions [Fig.\ref{fig:9}(a)]. With the boost from either FFA or MSI, small drusen could be identified as areas of hypofluorescence or hyperfluorescence in FFA or as regions with high reflectance in MSI. This results in a prominent shrinkage of the activated regions in the corresponding heatmap, e.g., the one that represents combined CFP and FFA. When saliency map features are further incorporated, these heating regions become even more compact. Although saliency maps do not directly target drusen in cases such as wet AMD, they capture related patterns, including hemorrhagic changes caused by drusen-related neovascularization or other vascular abnormalities. For PT \#2, optic disc pallor, characterized by a pale appearance of the optic disc, is marginally detectable using CFP. While FFA is not specifically designed to highlight pallor, it may indirectly reveal vascular abnormalities indicative of optic neuropathy. In contrast, MSI is particularly effective at detecting and quantifying pallor by analyzing wavelength-dependent reflectance patterns. As observed in Fig.\ref{fig:9}(a), MSI fundamentally enhances both localization and compactness of the heatmaps. Since saliency maps have already learned to identify the optic disc region, their inclusion further refines the heatmap focus. PT \#3 holds the most significant improvement in attention focus as the number of modalities increases. This could be explained by the presence of microhemorrhage \& microaneurysms, which are highly sensitive to detection with FFA and highly discriminative with MSI due to their strong absorption in the visible-to-near-infrared spectrum. As the saliency maps have been specifically trained to segment such lesions, the final heatmaps accurately highlight the latent abnormal regions, aligning closely with the ophthalmologists’ annotations. For PT \#4, heatmaps based on CFP alone show widespread activation, despite lesions being more readily identifiable across all the modalities when compared to PT \#3. This broader activation is partly due to the additional exudates characteristic of severe NPDR. With the incorporation of multimodal inputs, the model's focus becomes substantially more precise and lesion-specific, even for abnormalities occupying limited regions (as marked by the yellow arrows). These findings underscore the value of multimodal inputs in improving interpretability and lesion localization.

\begin{figure}[H]
    \centering
     \includegraphics[width=1.0\linewidth]{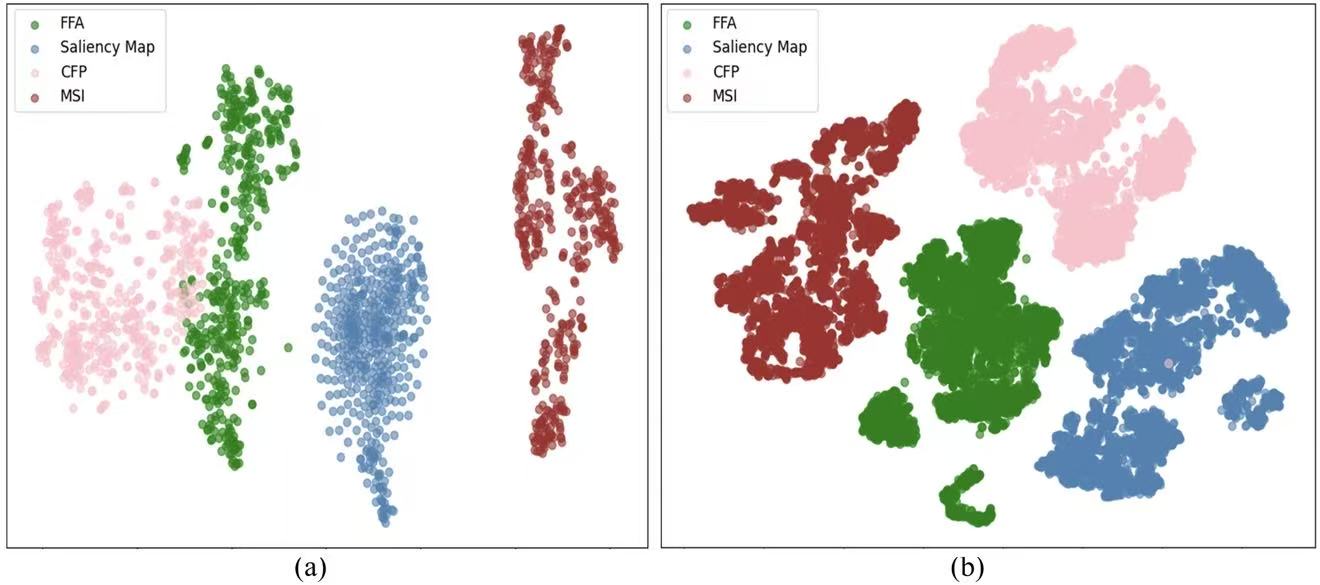} 
    \caption{Visualizations of feature embedding casted for (a) multi-label classification and (b) DR grading, to illustrate the independent contributions of each modality.}
    \label{fig:10}
\end{figure}

   To quantify the contribution of each modality, SHAP analysis was conducted [Fig.\ref{fig:9}(b)]. As expected, FFA exhibits the highest mean absolute SHAP value, consistent with its established role in vascular pathology diagnosis. Although all modalities are synthesized from the original CFP, the CFP still plays a central role due to its preservation of key structural and pathological features. Derived modalities, while complementary, inherently depend on the CFP content. Saliency maps contribute least according to SHAP scores, yet their inclusion improved spatial focus in the heatmaps as shown in Fig.\ref{fig:9}(a), demonstrating their value in fine-grained lesion localization.
   
   Feature embeddings of the four modalities are visualized using t-SNE in Fig.\ref{fig:10}, for both multi-label classification and DR grading. To focus on synthesis quality, these embeddings are extracted prior to CPAM-based feature calibration. Intuitively, the more realistic the synthesized data, the greater the independence among the modalities, as each captures distinct pathophysiological characteristics. This trend is basically reflected in Fig.\ref{fig:10}, where embeddings from intra-modality cluster tightly, and those from inter-modality keep distance. Slight overlapps exist between CFP and FFA [Fig.\ref{fig:10}(a)], probably due to shared visualization of retinal structures such as blood vessels, the optic disc, and macula. In contrast, MSI emphasizes biochemical and structural changes that may not be detectable with either CFP or FFA. Regarding the saliency maps, the segmented regions of interest exhibit varying degrees of representation across modalities, leading to more dispersed embeddings.
	
\begin{figure}[H]
    \centering
     \includegraphics[width=1.0 \linewidth]{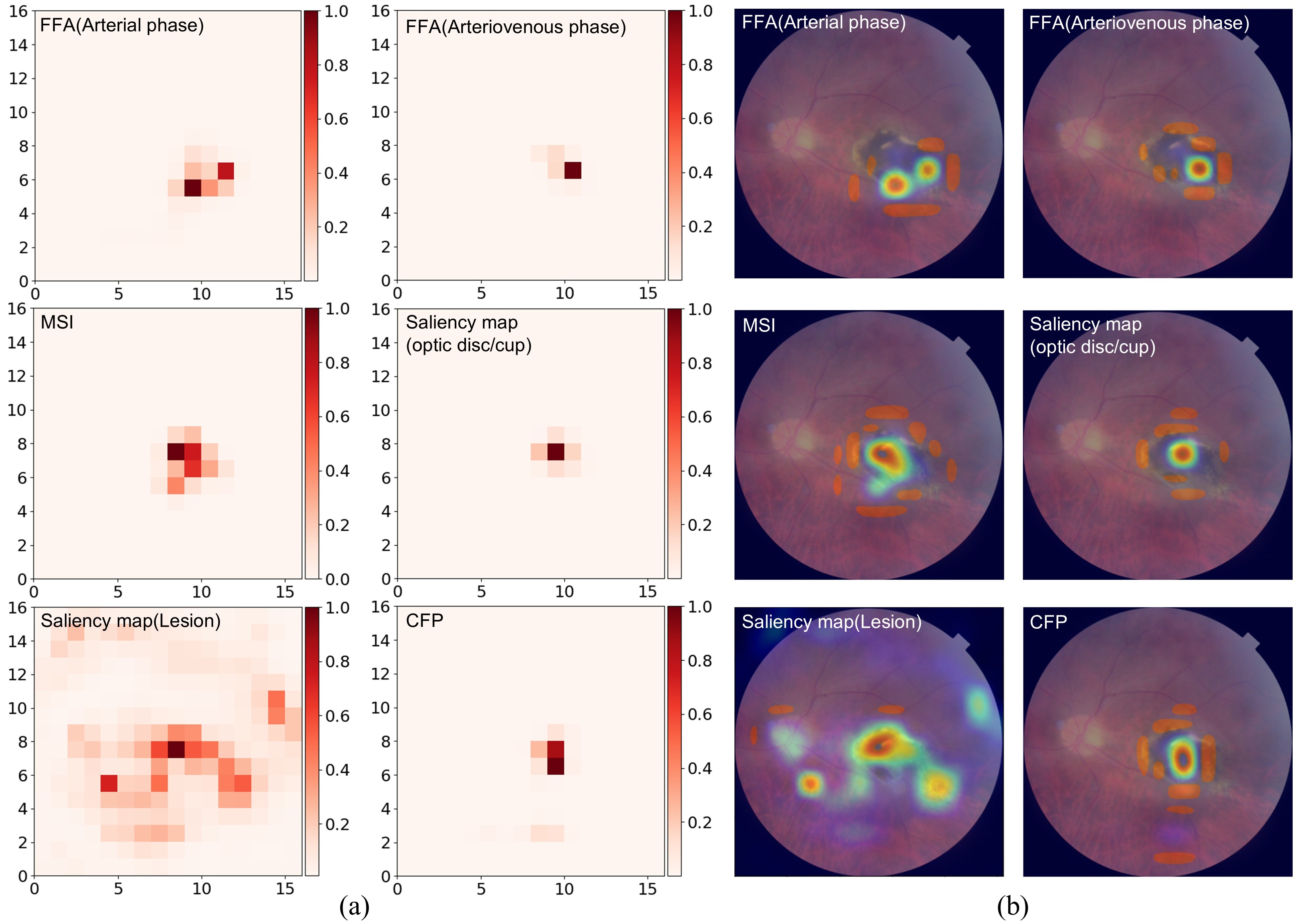} 
    \caption{Visualization of the results from the adaptive calibration mechanism applied to ARMD diagnosis: (a) attention maps learned for each modality, and (b) their overlay on the corresponding fundus images.}
    \label{fig:11}
\end{figure}

    In the proposed method, multimodal features are adaptively calibrated via CPAM according to downstream task. This step is pivotal in addressing common challenges in multimodal diagnosis, such as information redundancy and the need for flexible information integration. To assess the effectiveness of the calibration mechanism, attention maps for each modality are visualized in Fig.\ref{fig:11}. These attention maps are obtained from Eq.(4) by averaging the weights across all queries within the original matrix of dimension $d_k \times d_k$ , for a randomly selected case of ARMD. For better understanding, the original attention maps are upsampled to match the resolution of input image using cubic spline interpolation and overlaid onto the corresponding source images as presented in Fig.\ref{fig:11}(b). As illustrated, the attention varies fundamentally across modalities in both intensity and spatial distribution, although all maps consistently emphasize lesion areas. Notably, MSI provides the strongest attention, arguably due to its particular sensitivity to ARMD-related features \cite{Ma2023}. 

\subsection{Model robustness}

\begin{table}[ht]
\centering
\caption{Stability evaluation under different perturbations}
\resizebox{0.9\textwidth}{!}{
\begin{tabular}{|l|cccc|ccc|}
\hline
\textbf{Perturbations} & \multicolumn{4}{c|}{\textbf{Multi-label classification}} & \multicolumn{3}{c|}{\textbf{DR grading}} \\
\cline{2-8}
 & Precision & F1 & AUC & mAP & Kappa & Precision & Accuracy \\
\hline
None & 0.837 & 0.683 & 0.953 & 0.757 & 0.861 & 0.835 & 0.842 \\
Blur & 0.811 & 0.652 & 0.950 & 0.749 & 0.842 & 0.828 & 0.831 \\
Sharp & 0.816 & 0.648 & 0.951 & 0.741 & 0.833 & 0.821 & 0.825 \\
Distortion & 0.823 & 0.665 & 0.949 & 0.739 & 0.838 & 0.826 & 0.826 \\
\hline
\end{tabular}
}
\label{tab:stability_perturbations}
\end{table}

Model robustness was assessed by simulating common perturbations encountered in clinical practice, such as out-of-focus fundus imaging or cataract presence (blur), pupil dilation (sharpness), and vascular deformation (distortion). Blur was applied using a 5×5 Gaussian kernel with a standard deviation of 1; sharpening was performed using a standard 3×3 kernel; and distortion was simulated through a combination of random shearing, scaling, translation, and rotation operations. Table \ref{tab:stability_perturbations} summarizes the performance under these perturbations. For multi-label classification, the average performance drops are  2.25\%,\ 2.48\%\ and 1.77\%,\ for blur, sharpening and distortion, respectively. For DR grading, the corresponding drops are 1.45\%,\ 2.31\%,\ and 1.88\%.\ The most pronounced degradation occurs under sharpening perturbation, with reductions of 1.67\%\ in precision, 2.01\%\ in accuracy, and 3.25\%\ in kappa score. These results demonstrate that the proposed model maintains strong robustness against a variety of clinically relevant perturbations.

\section{Discussion}

Multimodal data substantially enhances diagnosis, prognosis, treatment planning, and patient monitoring by providing complementary clinical insights \cite{Li2020a} \cite{Foo2023}. However, the physical deployment of multimodal diagnostic systems suffers from both internal and external limitations as abovementioned \cite{Dalmaz2022}\cite{Zhang2022}. As such, a quasi-multimodal model is proposed for retinal disease classification and grading, which departs from the conventional synthesis–then–diagnosis paradigm in three key ways: i) modality-specific encoders are jointly finetuned with diagnostic supervision, directly aligning representation learning with clinical objectives instead of relying on synthesis fidelity as an indirect surrogate; ii) the proposed CPAM explicitly models pathophysiological correspondences at multiple scales, shifting multimodal integration from generic feature aggregation to pathology-aware calibration and alignment; iii) missing modalities are imputed at the representation level, propagating high-level, task-adapted features across stages. This strategy facilitates robustness by reducing the mismatch between pixel-level synthesis complexity and the relative simplicity of diagnostic prediction \cite{Praveen2024}\cite{Sun2024}, establishing a more direct pathway from incomplete modalities to reliable clinical outcomes.

The goal of multimodal synthesis is not to replicate visible content, but to reorganize, amplify, and disentangle diagnostically relevant features latent in the CFP, revealing them from multiple pathophysiological perspectives. This can be intuitively likened to color space transformations, e.g., converting RGB images into HSV or Lab spaces. While the underlying image content remains unchanged, these transformations yield alternative representations that emphasize specific visual features (e.g., luminance or chromaticity), making them more informative for downstream tasks like segmentation or classification \cite{Oukil2022}\cite{Mu2023}. 

In our framework, each synthesized modality contributes unique diagnostic value. Arterial-phase FFA, for instance, reveals arterial perfusion and initial signs of fluorescein leakage, facilitating the early-stage detection of relevant pathologies \cite{Moosavi2006}\cite{Wang2019}. MSI, on the other hand, enhances tissue discriminability by exploiting hybrid spectral features, particularly in the near-infrared range, which penetrates deeper layers and provides better visualization of choroidal structures. When combined with the synthesized arteriovenous-phase FFA, the model gains comprehensive insights into blood flow dynamics, vascular leakage \&\ permeability, temporal response and so on, enhancing the identification of subtle abnormalities such as non-perfused areas and pathological vascular structures \cite{Ying2008}\cite{Shen2024}. As demonstrated in Table \ref{tab:per_class_precision}, classification accuracy improves notably by 8.07\%\ (BRVO), 1.69\%\ (CRVO) and 35.41\%\ (CNV). The synthesized MSI data, spanning visible to near-infrared wavelengths, allows for high tissue penetration and thereby enhances diagnostic performance, especially for choroidal pathologies, achieving precision gains of 18.42\%\ (TSLN), 45.40\%\ (DN), and 29.41\%\ (MYA). Detection of ASR remains moderate, reflecting the inherent difficulty of this condition, for which other approaches report near-zero precision. ROC analysis and confusion matrices (Fig.\ref{fig:6}) further validate the superior performance in DR grading, with AUC improvements of 2.12\%\ (grade 0), 7.89\%\ (grade 1), 3.44\%\ (grade 2), and 5.68\%\ (grade 3). 

To further improve the image synthesis from a diagnostic perspective, the encoders responsible for multimodal representations are jointly finetuned by task-specific objectives and fidelity regularization. This encourages the encoders to learn features that are both modality-consistent and diagnostically discriminative \cite{Pan2021}\cite{Gao2023}\cite{Wang2024}. This approach could also enhance modality-specific encoder representational capacity by leveraging additional clinical data. Without finetuning, such improvements would require intensive training in Stage I with carefully registered image pairs, which is time-consuming and labor-intensive (see Appendix A, Section 1). In contrast, finetuning employs advanced loss functions that support unsupervised or self-supervised learning, effectively eliminating the need for strict registration. Moreover, incorporating diverse clinical data further exposes the model to potential external perturbations, enhancing generalizability and stability. As demonstrated in Section 4.5, performance changes by less than 3.25\%\ when diverse perturbations are applied to input CFP images.

Many challenges in multimodal diagnosis, such as data heterogeneity, spatial-temporal alignment and registration, can be naturally addressed in an all-in-one mode, where all modalities are synthesized from a single source. Nevertheless, issues regarding information redundancy and flexible feature integration remain critical. This is critical for multimodal diagnosis. For instance, although multimodal data improves multi-label classification (Table ]\ref{tab:ablation_modality}), coarse calibration (e.g., MFFM alone) limits the gains (Table \ref{tab:cpam_calibration_nodes}). To address this issue, CPAM is proposed to calibrate both modal-specific and modal-dependent features. By leveraging attention mechanisms, CPAM enables hierarchical and adaptive prioritization of discriminative features according to task-specific objectives, in a recursive fashion. This design ensures not only the preservation of intra-modality information but also effective inter-modality feature interaction. Node-wise evaluations of CPAM (Table \ref{tab:cpam_calibration_nodes}) highlight substantial improvements in precision and mAP, underscoring its capacity to deliver superior classification accuracy and overall performance. Visualization studies further support these findings as shown in Fig.\ref{fig:8}, where the addition of MSFC or MDFC promotes more compact clustering of disease and DR grade embeddings.

    The necessity and importance of each modality have also been demonstrated in ablation study. As expected, model performance generally increases monotonically with the integration of additional modalities, as presented in Table \ref{tab:ablation_modality}. These enhancements are visually interpreted in Fig.\ref{fig:9}(a), where lesion areas from selected patients are accurately highlighted through the attention-generated heatmaps. Comparing to using CFP alone, the quasi-multimodal method highly advances both the precision in localizing lesions and the effective coverage of clinically relevant regions even for those with extreme areas.
    
Despite its promising performance, this method has limitations. First, dataset imbalance leads to increased false negatives, particularly for diseases that share similar manifestations with more prevalent conditions. Second, the arterial-phase FFA dataset contains timing inconsistencies, potentially hindering the synthesis of fine arterial structures. This could be attributed to the fact that the arterial phase lasts only about a dozen seconds post-injection. Standardizing the FFA imaging protocol, particularly with regard to the timing of image acquisition, is a future direction. Third, involving additional modalities that target diverse diagnostic attentions could benefit, e.g., ophthalmoscope, indocyanine green angiography and so on. Fourth, adopting advanced generative models may improve both synthesis quality and the expressiveness of modality-specific features. Finally, external or multi-institutional validation was not performed in this study. One contributing factor is the scarcity of MSI data, as multispectral imaging has not been widely adopted in most hospitals or ophthalmic centers. Finally, external validation for the multi-label classification task has not yet been performed. To address this, we are initiating collaborations with multiple medical centers to obtain additional datasets covering all 19 disease categories (except the “other diseases” group).

\section{Conclusion}
This study introduces a unified learning framework that comprises quasi-multimodal data synthesis and fusion for retinal disease classification and grading. All multimodal data are synthesized solely from fundus images, significantly advancing automated fundus-based screening and diagnosis, and fostering a scalable diagnostic paradigm both within and beyond ophthalmic practice. To ensure alignment between the synthesized representations and diagnostic goals, the image generation process is jointly finetuned with the fusion pipeline under combined guidance of task-specific objectives and fidelity regularization. Within the fusion pipeline, two specialized modules are designed for modal-specific and modal-dependent feature calibration, to tackle with the common challenges in multimodal data processing. Experiments conducted on multi-label classification and DR grading demonstrated the superiority of the proposed method against other state-of-the-art ones. Contributions from each modality as well as the necessity of each calibration node are also validated through comprehensive ablation studies. Future work will focus on incorporating more clinically available modalities, leveraging advanced generative techniques, and enabling multi-center data collaboration to further promote the applicability, generalizability, and robustness in real-world medical scenarios.

\vspace{1em}
\section*{Declaration of competing interest}
\addcontentsline{toc}{section}{Declaration of competing interest}

We declare that we have no financial and personal relationships with other people or organizations that can inappropriately influence our work, there is no professional or other personal interest of any nature or kind in any product, service and/or company that could be construed as influencing the position presented in, or the review of, the manuscript.

\vspace{1em}
\section*{Acknowledgments}
\addcontentsline{toc}{section}{Acknowledgments}

This research was supported in part by the National Natural Science Foundation of China (62175183), Key Research and Development Program of Zhejiang Province (2024C03070, 2023C01041), Tianjin Metrology Science and Technology Project (2024TJMT007, 2024TJMT032), and Tianjin Municipal Education Commission Scientific Research Project (2024ZXZD017).

\vspace{1em}
\section*{Data and code availability.}
\addcontentsline{toc}{section}{Data and code availability.}

Data underlying the results presented in this paper are not publicly available but may be obtained from the authors upon reasonable request. The code is publicly available at https://github.com/zlu866/QMP-RETNet.

\newpage
\bibliography{ref}
\newpage
\appendix
\section{Modality-specific Representation Learning}

\subsection{FFA image synthesization}

\begin{figure}[H]
    \centering
    \renewcommand{\thefigure}{A.1}
     \includegraphics[width=1.0\linewidth]{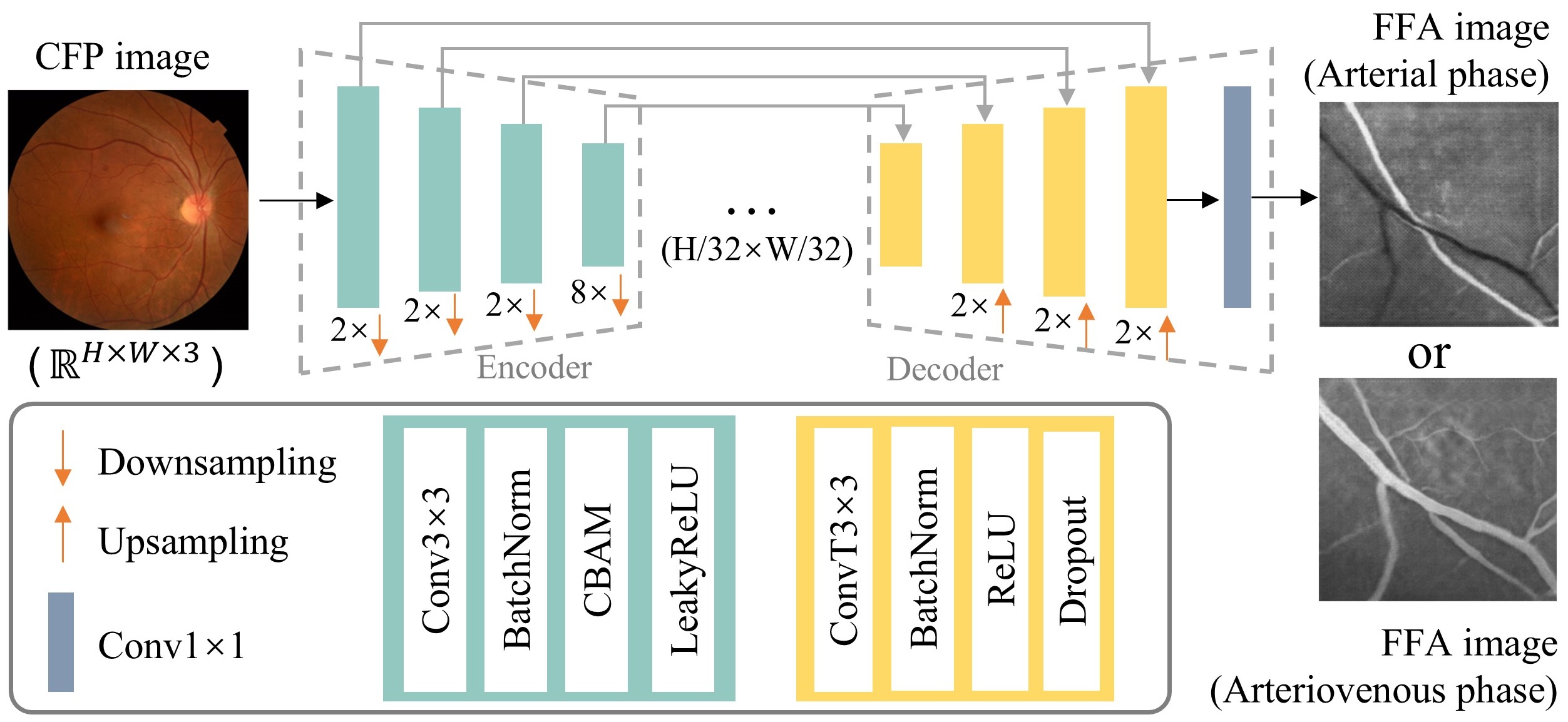} 
    \caption{Architecture of the model for FFA synthesization. The encoder and decoder components employed for multimodal diagnosis in Stage II are indicated.}
    \label{fig:A.1}
\end{figure}

FFA images in both arterial- and arteriovenous- phases are synthesized from CFP using a learning model backboned on the Pix2pix architecture \cite{Isola2017} . To enhance vascular feature extraction, Convolutional Block Attention Modules (CBAM) \cite{Woo2018} are integrated in each downsampling block as shown in Fig.\ref{fig:A.1}. By leveraging the combined advantages of contextual attention mechanism, a multi-scale discriminator, and an advanced adversarial loss, the model achieves promising FFA image generation. 

	Each data group includes one CFP image along with its corresponding arterial- and arteriovenous-phase FFA images. Prior to training, all image pairs were carefully aligned using a consistent two-stage image registration pipeline. First, coarse alignment was manually performed using BigWarp in ImageJ (version 1.54f), an interactive tool for landmark-based deformable registration. Then, fine-grained registration was performed automatically using a deep learning-based model specifically designed for multimodal retinal image registration \cite{Zhang2019}. For the CFP–FFA dataset, 80\%\ of the image pairs were randomly selected for training, and the remaining 20\%\ for testing. During training, each image pair was cropped into overlapping patches (128 × 128 pixels) using a sliding window with a stride of 128 pixels. The learning rate was initially set to 0.001 for the first 100 epochs and then gradually reduced to zero over the remaining 100 epochs via cosine annealing.
    
\begin{figure}[H]
\renewcommand{\thefigure}{A.2}
    \centering
     \includegraphics[width=1.0\linewidth]{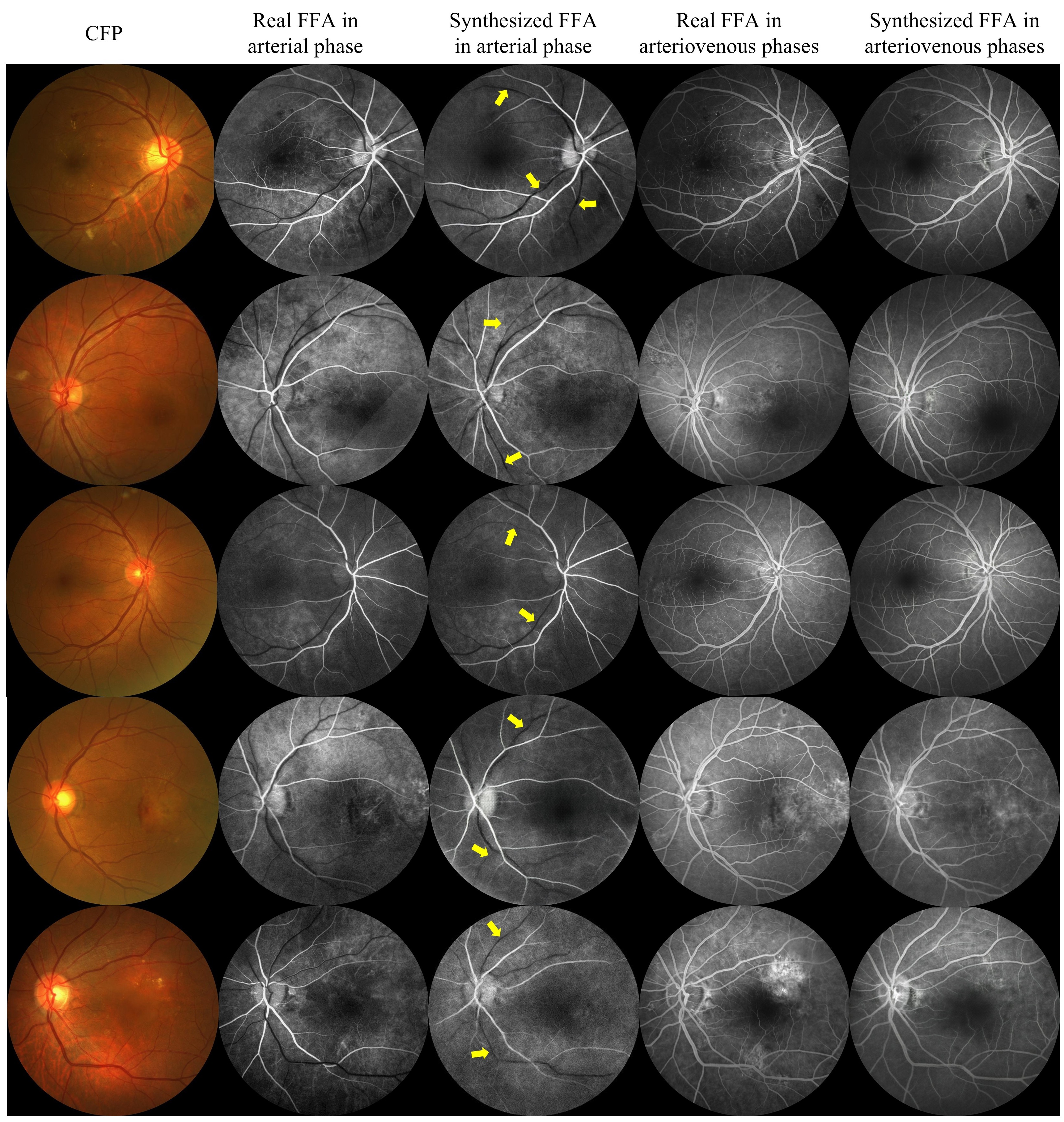} 
    \caption{Comparison between the synthesized FFA images and their real counterparts in both arterial and arteriovenous phases. In the synthesized arterial-phase FFA, veins are accurately and clearly depicted in black as labeled.}
    \label{fig:A.2}
\end{figure}

    Both quantitative and qualitative evaluations were conducted on an unseen dataset, with representative comparisons presented in Fig.\ref{fig:A.2}. To ensure fair comparison, the real FFA images were manually aligned with the corresponding CFP images. As shown in Fig.\ref{fig:A.2}, the synthesized arterial-phase images accurately depict veins in black, capturing fine-grained details such as small branches and filling patterns, as highlighted with yellow arrows. In the arteriovenous phase, veins appear bright in the synthesized images, closely resembling the true FFA contrast enhancement. This high-fidelity performance is largely attributed to the enforced pixel-level correspondence enabled by the accurate CFP–FFA alignment. Although unpaired image training could potentially be addressed using self-supervised learning methods that incorporate cycle-consistency regression, e.g., such as CycleGAN \cite{Zhu2017} and StarGAN \cite{Choi2018}, these approaches generally underperform in this context (not shown here). This is primarily due to the extreme imbalance between the sparse vascular structures and the overall fundus region. Compared to the arterial-phase FFA, arteriovenous-phase FFA could be intuitively regarded as a form of contrast-enhanced CFP, making it relatively easier to synthesize. This may explain why previous studies have predominantly focused on the arteriovenous phase \cite{Li2019a}\cite{Li2019b}\cite{Li2020a}\cite{Li2020b}. 
    
   Table \ref{tab:ffa_eval} presents the quantitative results in terms of structural similarity index measure (SSIM), multi-scale structural similarity index measure (MS-SSIM), peak signal-to-noise ratio (PSNR), and learned perceptual image patch similarity (LPIPS). Compared to arteriovenous-phase results, arterial-phase synthesis shows moderately reduced performance, with decreases of 25.17\%,\ 11.97\%,\ 4.55\%,\ and 34.04\%\ in SSIM, MS-SSIM, PSNR, and LPIPS, respectively. In addition to minor misalignment between the reference and synthesized images, a key contributing factor may be the latent inconsistency within the arterial-phase training dataset. This inconsistency stems from variations in the temporal delay between fluorescein injection and image acquisition, which can differ across individuals under the current imaging protocol.

\begin{table}[ht]
\centering
\renewcommand{\thetable}{A.1}
\caption{Quantitative evaluations on the synthesized FFA in both arterial and arteriovenous phases.}
\label{tab:ffa_eval}
\resizebox{0.6\linewidth}{!}{
\begin{tabular}{lcccc}
\toprule
\textbf{Phase} & \textbf{SSIM} & \textbf{MS-SSIM} & \textbf{PSNR} & \textbf{LPIPS} \\
\midrule
Arterial & 0.5312 & 0.6742 & 20.4428 & 0.3394 \\
Arteriovenous & 0.7099 & 0.7659 & 21.4192 & 0.2532 \\
\bottomrule
\end{tabular}
}
\end{table}

\subsection{MSI image synthesization}

Multispectral imaging (MSI) data spanning from visible to near-infrared wavelengths were synthesized solely from CFP images. Specifically, the interrogated wavelengths ($\lambda$) include 550 nm, 580 nm, 590 nm, 620 nm, 660 nm, 685 nm, and 740 nm.

\begin{figure}[H]
\renewcommand{\thefigure}{A.3}
    \centering
     \includegraphics[width=1.0\linewidth]{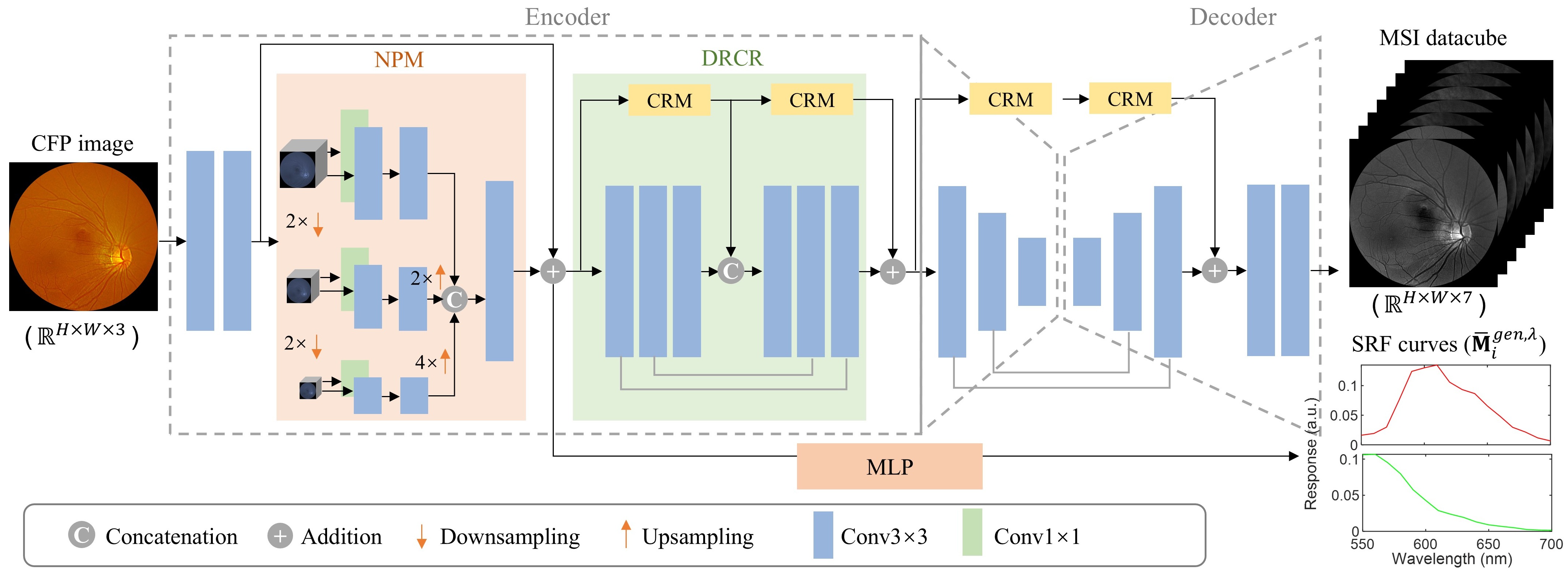} 
    \caption{Architecture of the model for MSI synthesization. A U-shaped structure is integrated with the Dense Residual Channel Recalibration (DRCR) block to produce the MSI datacube. Spectral response function (SRF) curves are generated through a Multi-Layer Perceptron (MLP) appended to the Non-local Purification Module (NPM). The encoder and decoder components used for multimodal diagnosis in Stage II are also indicated.}
    \label{fig:A.3}
\end{figure}

    The CFP-to-MSI mapping is built upon a state-of-the-art network, DRCR-Net, as illustrated in Fig.\ref{fig:A.3} \cite{Li2022}. The network incorporates a Non-local Purification Module (NPM) to mitigate latent perturbations at various scales as listed in Table \ref{tab:stability_perturbations}, and a Dense Residual Channel Recalibration (DRCR) block to enhance deep spatial-spectral feature extraction and intermediate feature interaction. Subsequently, a U-shaped architecture is employed to support hierarchical feature learning and balanced modeling of both high- and low- frequency components. Each DRCR block contains dual symmetrical Channel Recalibration Modules (CRMs), which recalibrate channel-wise feature responses by explicitly modeling inter-channel dependencies. Besides, a typical Multi-Layer Perceptron (MLP) appended to NPM generates spectral response functions (SRFs) from the input CFP image. To accommodate the specific configurations of CFP and MSI (e.g., number of channels and wavelengths), the network architecture was adapted with 2 input channels, 7 output channels, 21 intermediate feature channels, and 1 DRCR block.
    
   The CFP–MSI registration follows a workflow similar to that used for CFP–FFA registration. The key difference is that the CFP images are synthetically generated from the MSI datacube using Eq.(A.5). Consequently, the registration task shifts from an inter-modality (i.e., CFP vs. MSI) to an intra-modality problem (within MSI only). In practice, all spectral bands within each synthesized MSI datacube were aligned to the image at 550 nm. The training/test group split, cropping strategy, and learning rate schedule were identical to those used for FFA image synthesis.

Assuming that the target emission has a spectral power distribution $\mathcal{S}^{(\lambda)}$ discretized at the above wavelengths, and the SRFs of the sensor inside a CFP instrument for the red, green, and blue channels are $\mathbf{M}_R^{(\lambda)},\ \mathbf{M}_G^{(\lambda)},\ \text{and}\ \mathbf{M}_B^{(\lambda)}$ ,respectively. The measurement of the i-th channel at pixel location (h,w) can be formulated as

\begin{equation}
    c_i(h,w) = \sum_{\lambda} S^{(\lambda)}(h,w) M_i^{(\lambda)}, \quad i \in \{R,G,B\}.
    \tag{A.1}
\end{equation}

Since the effective band of the MSI instrument (RHA2020, Annidis Corporation) starts at 550nm, the MSI datacube is synthesized by excluding the blue channel of CFP.

\begin{figure}[H]
\renewcommand{\thefigure}{A.4}
    \centering
     \includegraphics[width=1.0\linewidth]{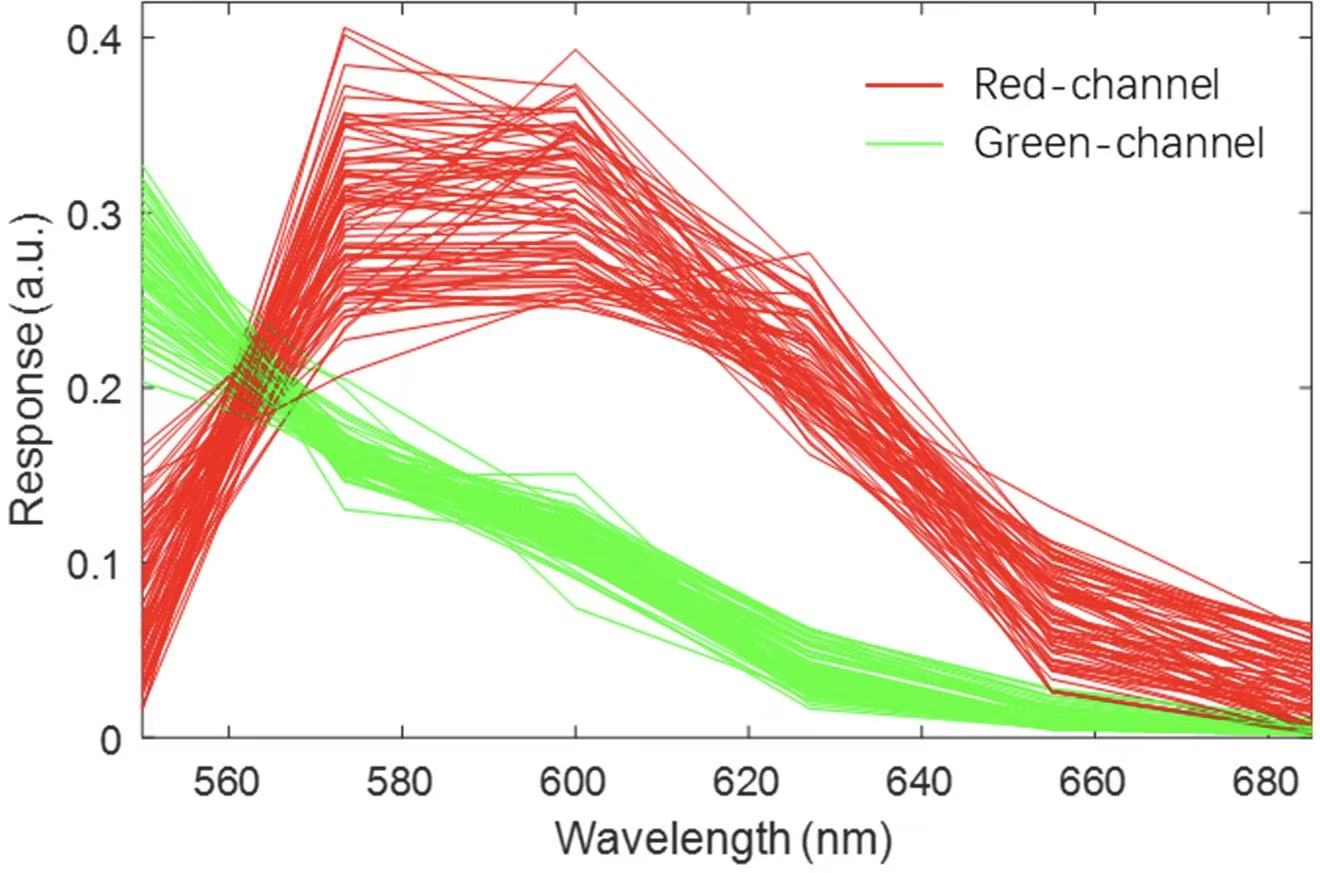} 
    \caption{Examples of perturbed spectral response functions (SRFs) employed for generating the training dataset. The baseline SRF curve corresponds to the sensor response of the MSI instrument employed in this study.}
    \label{fig:A.4}
\end{figure}

    Since imaging sensors inside MSI instruments could vary significantly, and their SRFs are typically difficult to precisely measure, a random spectral mapping strategy was adopted during training stage to enhance adaptability and generalizability. Specifically, the SRF was generated by perturbing the baseline distribution calibrated on the MSI instrument used in this study. The perturbation follows

\begin{equation}
    \overline{M}_i^{(\lambda)} = (1 + \Phi_A) \odot (M_i^{(\lambda)} \otimes \Phi_S),
    \tag{A.2}
\end{equation}

where $\Phi_A$ follows a standard normal distribution with the same dimension as $M_i^{(\lambda)}$ ; $\Phi_S$ also follows a standard normal distribution with its dimension randomly set between 1/15 to 1/3 of that of $M_i^{(\lambda)}$, $\odot$ denotes the Hadamard product; and $\otimes$ is the convolution operator. This formulation allows $\Phi_A$ and $\Phi_S$ to independently modulate the amplitude and spectral coverage of the SRF, respectively. Examples of perturbed SRFs used to generate the training dataset are illustrated in Fig.\ref{fig:A.4}. To enhance convergence and training stability, auxiliary supervision on SRF is introduced

\begin{equation}
    \mathcal{L}_{\text{MSE}} = \left\| \overline{\mathbf{M}}_i^{(\lambda)} - \overline{\mathbf{M}}_i^{gen,\lambda} \right\|_2^2,
     \tag{A.3}
\end{equation}

where $\overline{\mathbf{M}}_i^{gen,\lambda}$ is the SRF generated by the network. Besides, a cycle-consistency loss is employed to regularize the reconstructed CFP image, expressed as

\begin{equation}
    \mathcal{L}_{cyc} = \frac{1}{HW} \sum_i \sum_{h,w} \left| \mathbf{C}_i(h,w) - \mathbf{C}_i'(h,w) \right| \odot \mathbf{C}_i(h,w),
    \tag{A.4}
\end{equation}

where $\odot$ is the Hadamard division; $\mathbf{C}_i$ and $\mathbf{C}_i'$ represent the original and reconstructed CFP images, respectively, defined as

\begin{equation} \tag{A.5}
\left\{
\begin{aligned}
\mathbf{C}_i(h,w) &= \sum_{\lambda} \mathbf{S}^{(\lambda)}(h,w) \, \overline{\mathbf{M}}_i^{(\lambda)} \\
\mathbf{C}_i'(h,w) &= \sum_{\lambda} \mathbf{S}'^{(\lambda)}(h,w) \, \overline{\mathbf{M}}_i^{gen,\lambda}
\end{aligned}
\right.,
\end{equation}

with $\mathbf{S}'^{(\lambda)}$ being the synthesized MSI datacube. To further enhance perceptual quality, the original loss in DRCR is replaced with the SSIM loss, formulated by

\begin{equation} \tag{A.6}
\mathcal{L}_{ssim} = \sum_{\lambda} \left[
1 - \frac{\left( 2\mu_{\mathbf{S}^{(\lambda)}} \mu_{\mathbf{S}'^{(\lambda)}} + \xi_1 \right)
\left( 2\sigma_{\mathbf{S}^{(\lambda)} \mathbf{S}'^{(\lambda)}} + \xi_2 \right)}
{\left( \mu_{\mathbf{S}^{(\lambda)}}^2 + \mu_{\mathbf{S}'^{(\lambda)}}^2 + \xi_1 \right)
\left( \sigma_{\mathbf{S}^{(\lambda)}}^2 + \sigma_{\mathbf{S}'^{(\lambda)}}^2 + \xi_2 \right)}
\right],
\end{equation}

where $\mu$ and $\sigma$ represent the mean and standard deviation, respectively; $\sigma^2$ indicate the variance, respectively; $\xi_1$ and $\xi_2$ are small constants to avoid division by zero. The full objective is organized as

\begin{equation}
\tag{A.7}
    \mathcal{L}_{total} = \mathcal{L}_{cyc} + 0.5\mathcal{L}_{SSIM} + 0.5\mathcal{L}_{MSE}
\end{equation}

\begin{figure}[H]
\renewcommand{\thefigure}{A.5}
    \centering
     \includegraphics[width=1.0\linewidth]{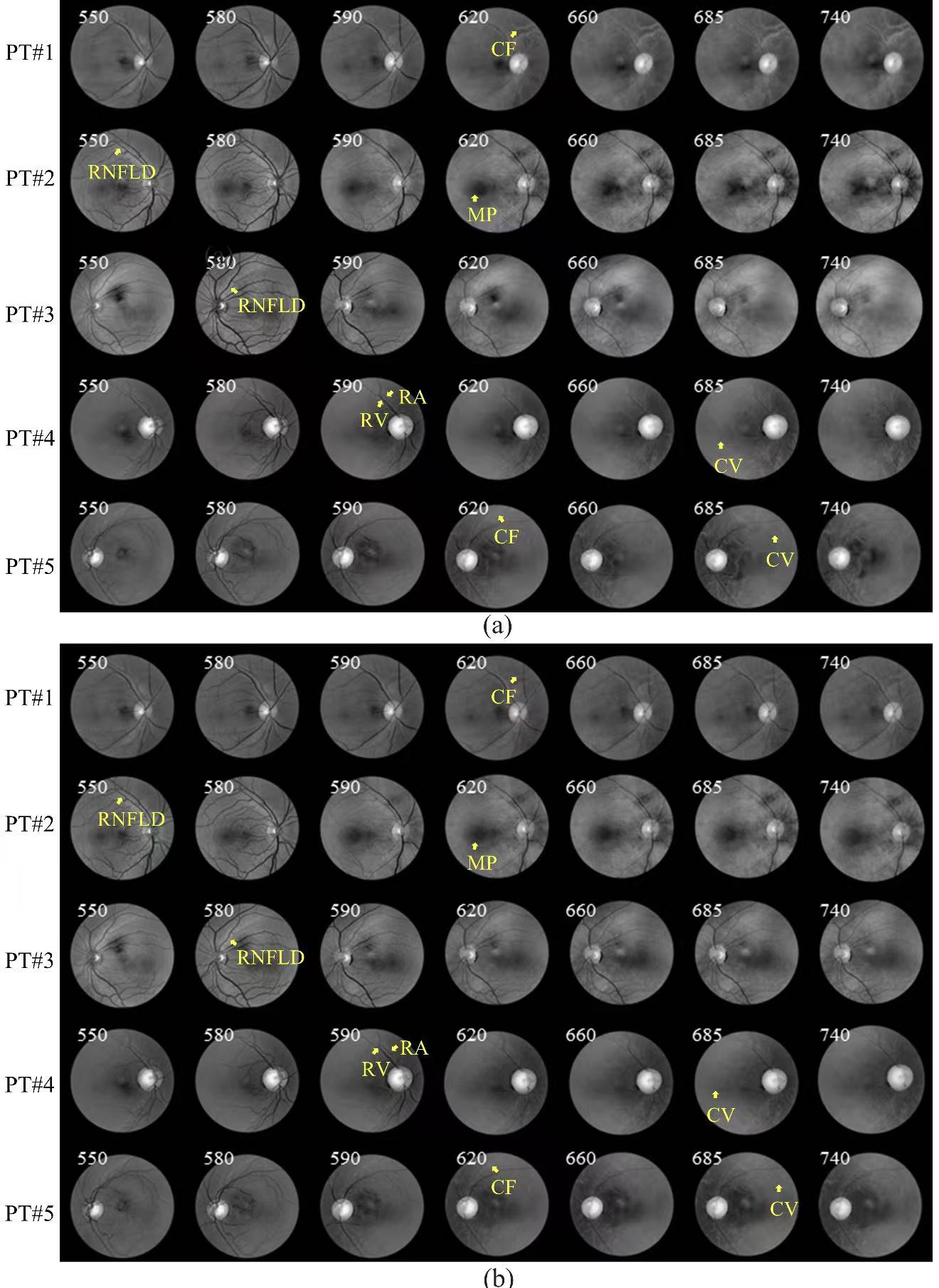} 
    \caption{Comparison between (a) real, and (b) synthesized MSI images across various wavelengths ranging from 550 nm to 740 nm for five representative cases. The yellow arrows highlight features including retinal nerve fiber layer defects (RNFLD), retinal arteries (RA), retinal veins (RV), macular pigment (MP), choroidal folds (CF), and choroidal vessels (CV).}
    \label{fig:A.5}
\end{figure}

The synthesized MSI data is exemplified in Fig.\ref{fig:A.5}, and quantitatively evaluated in Table \ref{tab:msi_eval} in terms of MS-SSIM, PSNR and LPIPS. For fair comparison, real MSI data were spatially aligned to the reference image at 550 nm. As observed from Fig.\ref{fig:A.5}, images at different wavelength exhibit distinct textures and features. This phenomenon arises from the fact that each wavelength corresponds to specific tissue absorption and scattering coefficients, leading to varying penetration depths and thus revealing different anatomical and pathological characteristics. For instance, localized retinal nerve fiber layer defects (RNFLD) appear as white striations posterior to the vascular arcades within the wavelengths between 550-590nm (e.g., PTs \#2 and \#3). Beyond this range, the contrast of retinal arteries (RA) gradually diminishes, while retinal veins (RV) remain visible. Above 600nm, macular pigment (MP) deposition becomes clearly discernible (e.g., PT \#2), and choroidal folds (CF) emerge as alternating hyper- and hypo-reflective bands at the choroid-RPE interface (e.g., PTs \#1 and \#5). Choroidal vessels (CV) are best visualized beyond 680 nm (e.g., PTs \#4 and \#5). Given that the effective spectral response of the imaging sensor spans from 550 nm to 590 nm, the synthesized MSI images demonstrate higher accuracy within this wavelength range, as found in Table \ref{tab:msi_eval}.

\begin{table}[ht]
\renewcommand{\thetable}{A.2}
\centering
\caption{Quantitative evaluations on synthesized MSI datacube}
\label{tab:msi_eval}
\resizebox{0.6\linewidth}{!}{
\begin{tabular}{lccc}
\toprule
\textbf{Wavelength(nm)} & \textbf{MS-SSIM} & \textbf{PSNR (dB)} & \textbf{LPIPS} \\
\midrule
550  & 0.9730 & 35.28 & 0.1337 \\
580  & 0.9691 & 33.83 & 0.1248 \\
590  & 0.9684 & 34.09 & 0.1408 \\
620  & 0.9408 & 30.35 & 0.1662 \\
660  & 0.9236 & 29.02 & 0.1686 \\
685  & 0.9246 & 29.06 & 0.1697 \\
740  & 0.9227 & 27.56 & 0.1732 \\
\bottomrule
\end{tabular}
}
\end{table}

\subsection{Synthesization of Saliency map}
\begin{figure}[H]
\renewcommand{\thefigure}{A.6}
    \centering
     \includegraphics[width=1.0\linewidth]{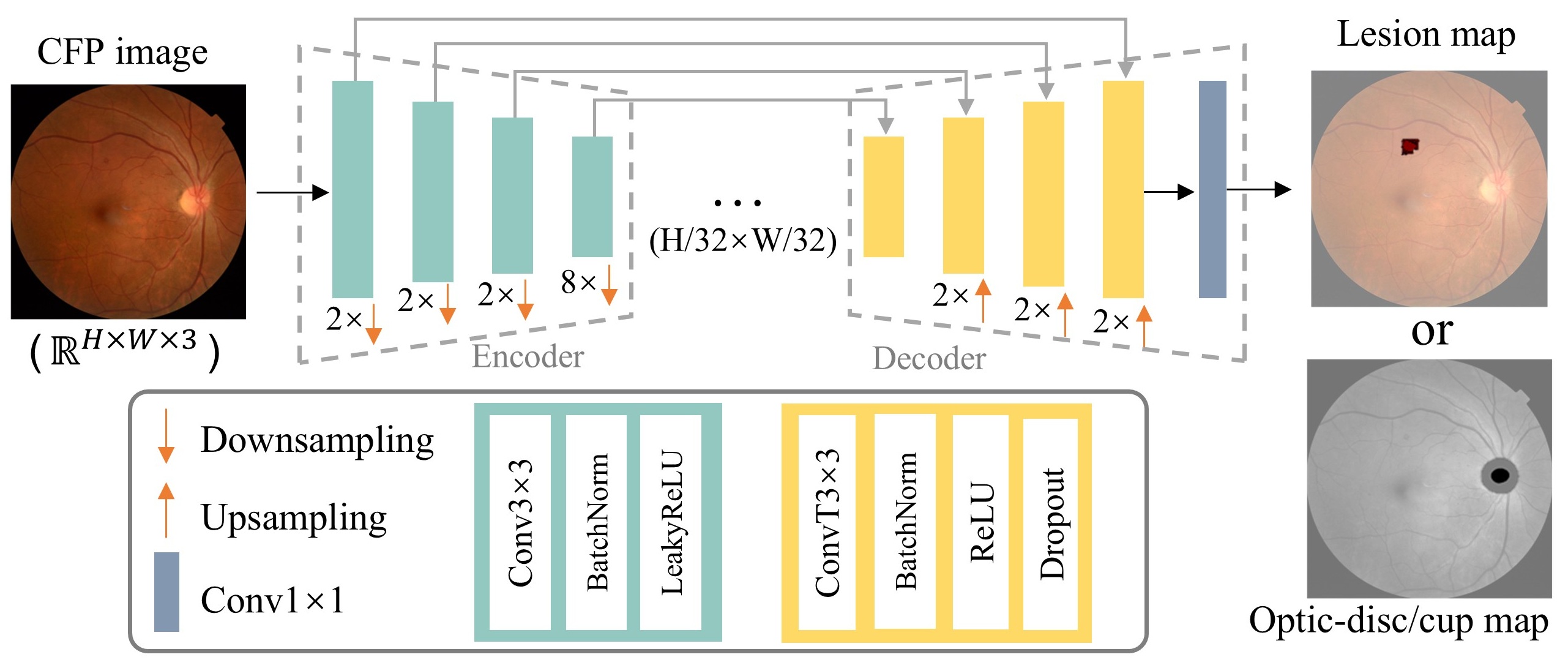} 
    \caption{Architecture of the model for saliency map synthesization. The encoder and decoder components employed for multimodal diagnosis in Stage II are indicated.}
    \label{fig:A.6}
\end{figure}

Saliency maps were synthesized to highlight both latent lesion regions and optic disc/cup areas within CFP images. These two types of regions were segmented using two independent networks, respectively, with each based on a U-Net backbone, an architecture widely recognized for its effectiveness in medical image segmentation. Given the fine structural details and large intra-class variations in terms of shape, size, color, and intensity, the original U-Net was improved by deepening the downsampling layers. This modification expands the receptive field, allowing the network to capture richer semantic information, as illustrated in Fig.\ref{fig:A.6}. Further details regarding the segmentation of lesions and optic disc/cup are provided separately below.

\subsubsection{Learning on lesion segmentation}
The lesions considered here include hard exudates (EX), soft exudates (SE), microaneurysms (MA) and haemorrhage (HE), which have been commonly found in diabetic retinopathy, age-related macular degeneration, and other diseases associated with microvascular rupture or leakage. The training/test group split followed the original protocol of the public dataset, while the cropping strategy and learning rate schedule were kept consistent with those used for FFA image synthesis.

The loss function involoves both the cross-entropy ($\mathcal{L}_{ce}$) loss and the dice loss ($\mathcal{L}_{dice}$), defined as

\begin{equation} \tag{A.8}
\left\{
\begin{aligned}
\mathcal{L}_{ce} &= -\frac{1}{HW} \sum_{h,w} \sum_n y_n(h,w) \ln\left[\tilde{y}_n(h,w)\right] \\
\mathcal{L}_{dice} &= 1 - \frac{1}{N} \sum_{n=1}^N \frac{2 \sum_{h,w} y_n(h,w) \tilde{y}_n(h,w) + \xi}{\sum_{h,w} \left[ y_n(h,w) + \tilde{y}_n(h,w) \right] + \xi}
\end{aligned}
\right.,
\end{equation}

where $\tilde{y}_n$ and $y_n$ are the predicted and ground-truth masks for the n-th label, respectively; N is the total of lesion types; and $\xi$ is a small constant used to prevent division by zero. The total loss is expressed as

\begin{equation} \tag{A.9}
\mathcal{L}_{total} = \mathcal{L}_{ce} + 0.4\mathcal{L}_{dice},
\end{equation}

\begin{figure}[H]
\renewcommand{\thefigure}{A.7}
    \centering
     \includegraphics[width=1.0\linewidth]{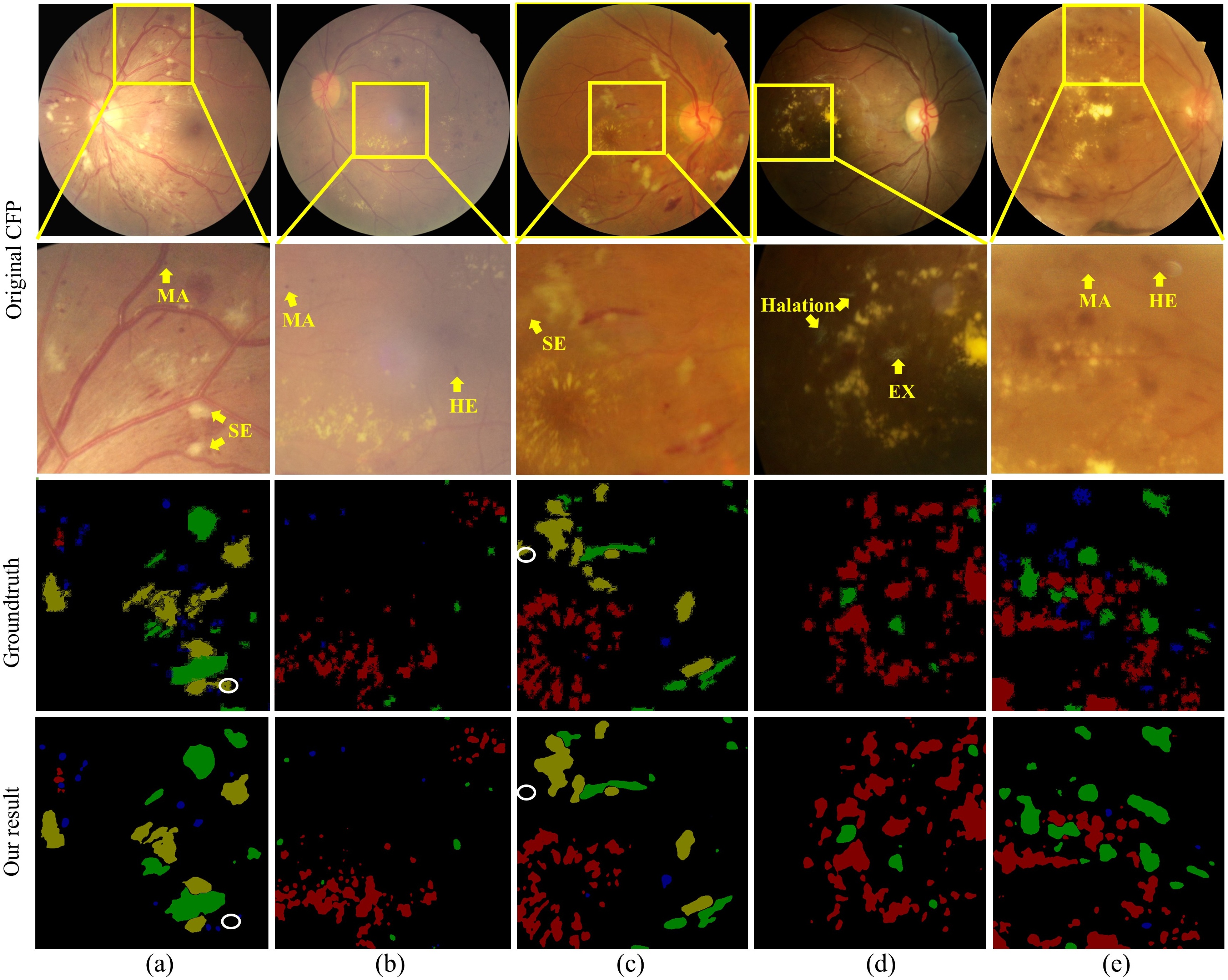} 
    \caption{Lesion segmentation results for five representative cases from (a) to (e), with red, green, yellow, and blue representing hard exudates (EX), soft exudates (SE), microaneurysms (MA) and haemorrhage (HE) lesions, respectively. Yellow arrows indicate lesion areas and halations. White circles mark SE segmentation failures, as labeled in both the ground truth and our results.}
    \label{fig:A.7}
\end{figure}

Experimental results were evaluated in terms of Dice and Intersection over Union (IoU). The results are presented in Table \ref{tab:appendix_lesion_segmentation}, and visualized in Fig.\ref{fig:A.7}. As shown, performance on MA and HE is relatively poorer among the four lesion types. Compared to EX, Dice and IoU scores for MA drop by 25.54\% and 28.79\%, respectively. The latent reasons are summarized as below. First, the extreme area of MA make it challenging to detect, e.g., the regions denoted by yellow arrows in Fig.\ref{fig:A.7}(a). In contrast, EX (yellow plaques) and SE (wool-like white or grayish-white regions) typically cover larger areas, facilitating their identification. Second, MA and small HE frequently share similar shapes and structures, as marked with yellow arrows in Fig.A.\ref{fig:A.7}(b) and (e). Additionally, SE lesions [Fig.\ref{fig:A.7}(a) and (c)] show lower contrast than EX [Fig.\ref{fig:A.7}(d)], leading to decreases of 7.66\% in Dice score and 10.84\% in IoU. In addition to lesion-specific characteristics, physiological reflections could significantly degrade segmentation accuracy. As illustrated in Fig.\ref{fig:A.7}(d), these reflections, which manifest as patchy halations or membrane-like glare, exhibit high morphological similarity to EX.

\begin{table}[ht]
\centering
\renewcommand{\thetable}{A.3}
\caption{Quantitative evaluations on lesion segmentation}
\label{tab:appendix_lesion_segmentation}
\begin{tabular*}{0.5\textwidth}{@{\extracolsep{\fill}} lcc @{}}
\toprule[1pt]
\textbf{Lesion type} & \textbf{Dice} & \textbf{IoU} \\
\midrule
EX & 0.8404 & 0.7535 \\
HE & 0.7112 & 0.6081 \\
SE & 0.7760 & 0.6718 \\
MA & 0.6257 & 0.5365 \\
\bottomrule[1pt]
\end{tabular*}
\renewcommand{\thetable}{\arabic{table}}
\end{table}

\subsubsection{Learning on optic disc/cup segmentation}
The Cup-to-Disc Ratio (CDR), defined as the ratio between diameters of optic cup and optic disc, is a critical metric for glaucoma screening. The segmentation of the optic cup and optic disc was performed using the same network architecture and loss functions as those applied for lesion segmentation. The training/test group split followed the original protocol of the public dataset, while the cropping strategy and learning rate schedule were kept consistent with those used for FFA image synthesis.

\begin{figure}[H]
\renewcommand{\thefigure}{A.8}
    \centering
     \includegraphics[width=1.0\linewidth]{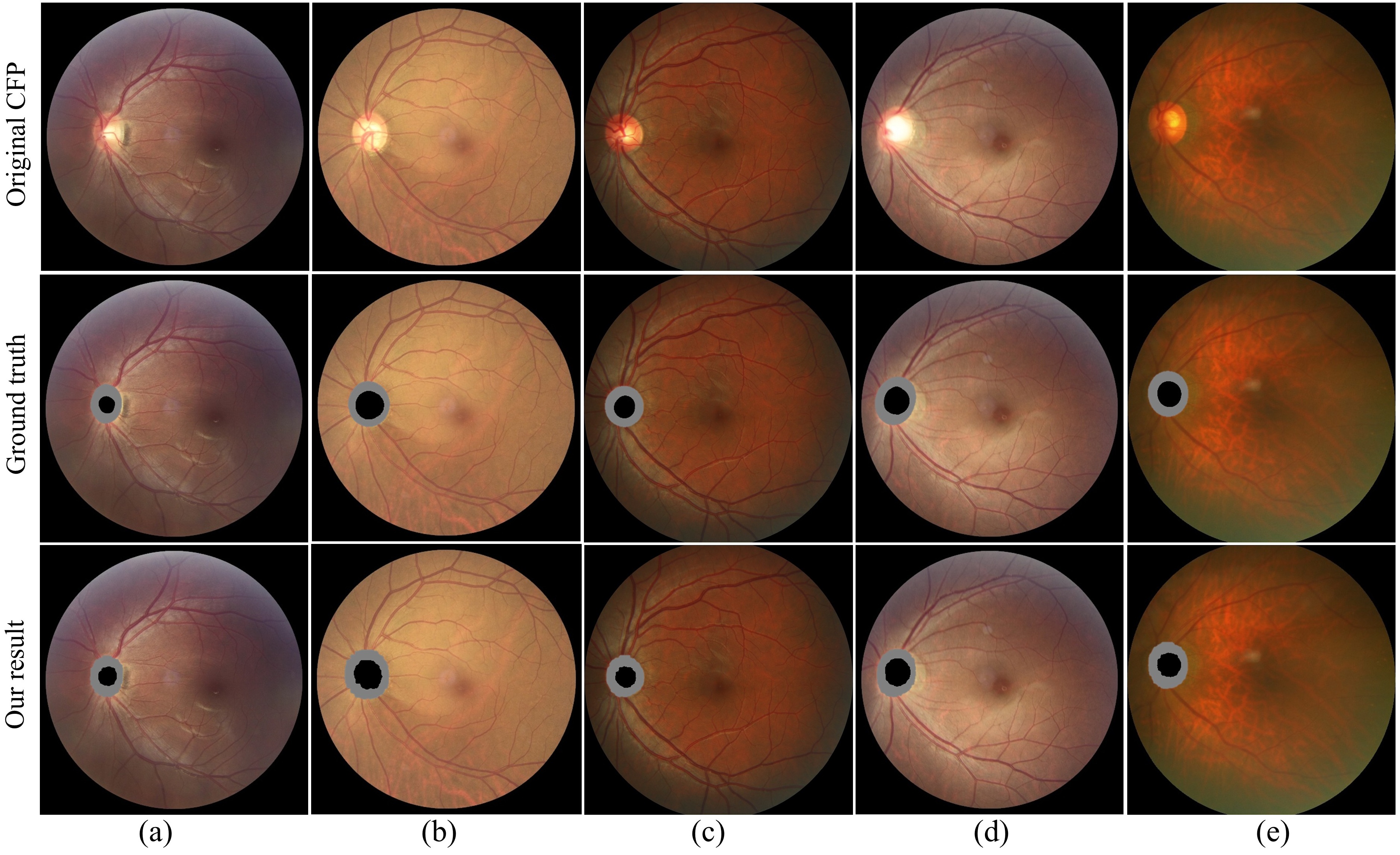} 
    \caption{Segmentation results of the optic disc/cup for five representative cases from (a) to (e). The optic disc and optic cup regions are filled with black and gray, respectively.}
    \label{fig:A.8}
\end{figure}

    As found in Fig.\ref{fig:A.8}, the boundaries of optic cup in our results exhibit pronounced jaggedness compared to the ground truth annotations. This observation is further confirmed by the quantitative results as presented in Table \ref{tab:appendix_disc_cup_segmentation}, where the Dice and IoU scores for the optic cup are 4.11\% and 14.83\% lower than those for the optic disc, respectively. The main contributing factor is the dense vascular network in the optic cup region, which frequently overlaps the structure and introduces segmentation ambiguity. Additionally, the limited spatial resolution may further compromise segmentation accuracy.

\begin{table}[ht]
\centering
\renewcommand{\thetable}{A.4}
\caption{Quantitative evaluations on optic disc/cup map segmentation}
\label{tab:appendix_disc_cup_segmentation}
\begin{tabular*}{0.4\textwidth}{@{\extracolsep{\fill}} lcc @{}}
\toprule[1pt]
\textbf{Target} & \textbf{Dice} & \textbf{IoU} \\
\midrule
disc & 0.9013 & 0.8716 \\
cup & 0.8642 & 0.7423 \\
\bottomrule[1pt]
\end{tabular*}
\renewcommand{\thetable}{\arabic{table}}
\end{table}



\newpage
\section{Additional Experimental Results}

\begin{table}[ht]
\centering
\renewcommand{\thetable}{B.1}
\caption{External validation results for DR grading on the EyePACS dataset}
\label{tab:appendix_b_dr_grading}
\begin{tabular*}{0.9\textwidth}{@{\extracolsep{\fill}} lccc @{}}
\toprule[1pt]
\textbf{Approach} & \textbf{Kappa} & \textbf{Precision} & \textbf{Accuracy} \\
\midrule
ResNet-18\,(He et al., 2016) & 0.595 & 0.621 & 0.571 \\
Inception-v3\,(Szegedy et al., 2016) & 0.619 & 0.641 & 0.568 \\
DenseNet-121\,(Huang et al., 2017) & 0.664 & 0.603 & 0.592 \\
RETFound\,(Zhou et al., 2023) & 0.650 & 0.684 & 0.660 \\
MedMamba\,(Yue and Li, 2024) & 0.668 & 0.691 & 0.674 \\
Ours & \textbf{0.705} & \textbf{0.718} & \textbf{0.701} \\
\bottomrule[1pt]
\end{tabular*}
\renewcommand{\thetable}{\arabic{table}}
\end{table}

\newpage

\begin{figure}[H]
\renewcommand{\thefigure}{B.1}
    \centering
     \includegraphics[width=1.0\linewidth]{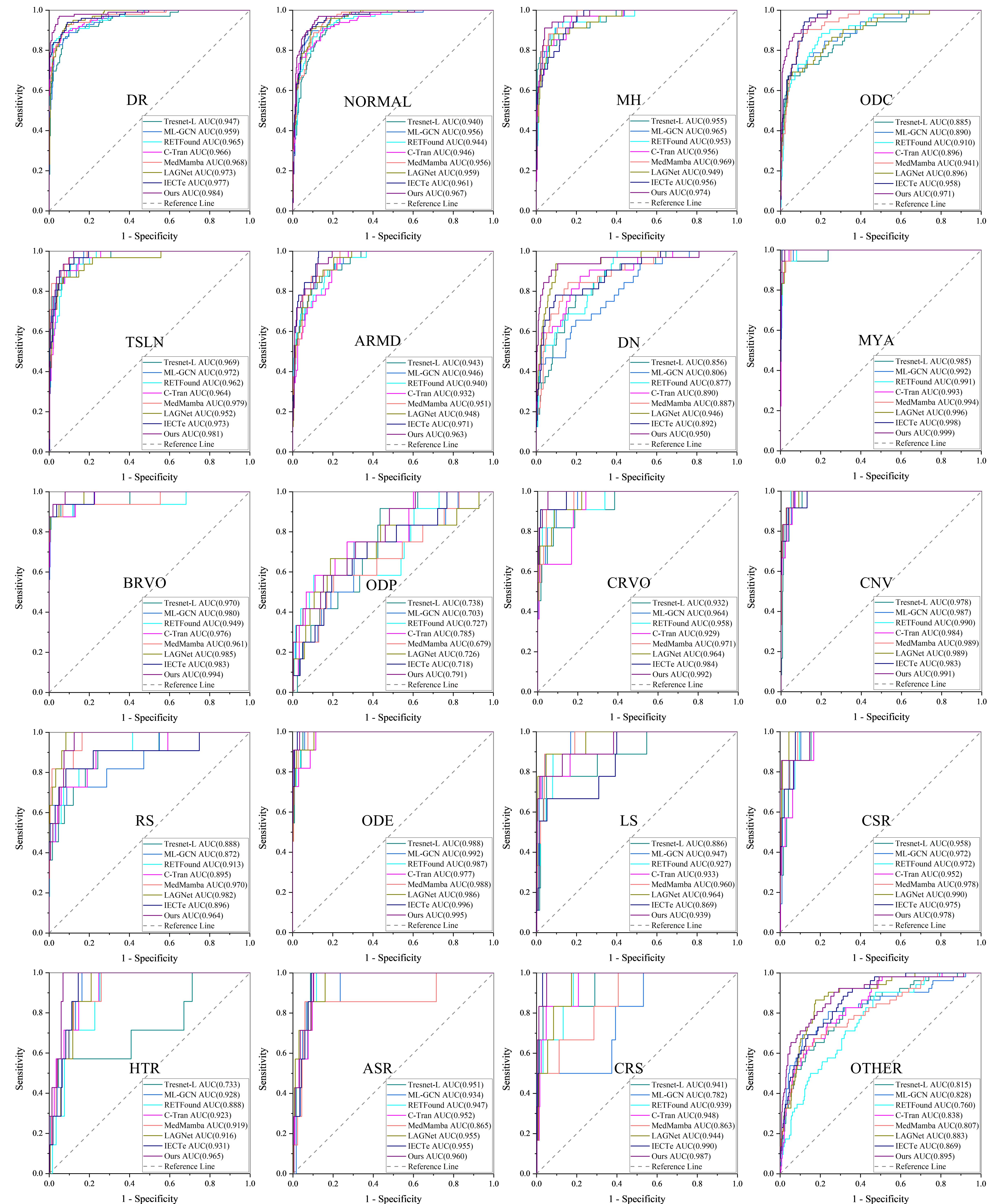}
    \caption{Comparison of ROC curves across 20 disease categories between our method and other approaches.}
    \label{fig:B.1}
\end{figure}

\end{document}